\documentclass[a4paper,fleqn]{cas-dc}

\usepackage[numbers]{natbib}
\usepackage{amsmath}
\usepackage{graphicx}
\usepackage[singlelinecheck=false]{caption}

\usepackage{paralist}
\usepackage{multirow}
\usepackage{makecell}
\usepackage{array}
\usepackage{booktabs}
\usepackage{slashbox}
\newcolumntype{P}[1]{>{\centering\arraybackslash}p{#1}}

\usepackage{amsthm}
\usepackage{subcaption}
\newcommand{\norm}[1]{\left\lVert#1\right\rVert}

\usepackage{amsmath}
\usepackage{graphicx}
\usepackage{textcomp}
\usepackage{parskip}
\usepackage{comment}
\usepackage{ragged2e}
\usepackage{amsfonts}

\usepackage{hyperref}
\hypersetup{
   colorlinks=true,
    linkcolor=blue,
    filecolor=magenta,      
    urlcolor=cyan,
}

\def\tsc#1{\csdef{#1}{\textsc{\lowercase{#1}}\xspace}}
\tsc{WGM}
\tsc{QE}
\tsc{EP}
\tsc{PMS}
\tsc{BEC}
\tsc{DE}

\newtheorem{theorem}{Theorem}
\newtheorem{lemma}{Lemma}
\newtheorem{assumption}{Assumption}
\newtheorem{property}{Property}
\newdefinition{rmk}{Remark}

\begin{document}
\let\WriteBookmarks\relax
\def\floatpagepagefraction{1}
\def\textpagefraction{.001}

\shorttitle{<short title of the paper for running head>}    

\shortauthors{<short author list for running head>}  

\title [mode = title]{AA-DLADMM: An Accelerated ADMM-based Framework for Training Deep Neural Networks}  



%

\author[1]{Zeinab Ebrahimi}
\author[2]{Gustavo Batista}
\author[1]{Mohammad Deghat}







 \affiliation[1]{organization={School of Mechanical and Manufacturing Engineering},
             addressline={ University of New South Wales}, 
             city={Sydney},
             postcode={NSW 2052}, 
             country={Australia}}

\affiliation[2]{organization={School of Computer Science and Engineering},
             addressline={ University of New South Wales}, 
             city={Sydney},
             postcode={NSW 2052}, 
             country={Australia}}










\begin{abstract}
Stochastic gradient descent (SGD) and its many variants are the widespread optimization algorithms for training deep neural networks. However, SGD suffers from inevitable drawbacks, including vanishing gradients, lack of theoretical guarantees, and substantial sensitivity to input. The Alternating Direction Method of Multipliers (ADMM) has been proposed to address these shortcomings as an effective alternative to the gradient-based methods. It has been successfully employed for training deep neural networks. However, ADMM-based optimizers have a slow convergence rate. This paper proposes an Anderson Acceleration for Deep Learning ADMM (AA-DLADMM) algorithm to tackle this drawback. The main intention of the AA-DLADMM algorithm is to employ Anderson acceleration to ADMM by considering it as a fixed-point iteration and attaining a nearly quadratic convergence rate. We verify the effectiveness and efficiency of the proposed AA-DLADMM algorithm by conducting extensive experiments on four benchmark datasets contrary to other state-of-the-art optimizers. 

\end{abstract}




\maketitle

\section{Introduction}\label{intro}


Stochastic gradient descent (SGD) and its many variants are well-known methods used as optimization algorithms for training deep neural networks, which apply the back-propagation method. However, multiple downsides of SGD pose a barrier to its outspread applications. For example, it suffers from a gradient vanishing phenomenon in which the error signal decreases as the gradient is backpropagated \cite{hanin2018neural}. In addition, SGD is immensely sensitive to poor conditioning, and therefore a slight change in the input would result in a drastic alteration in the gradient \cite{novak2018sensitivity}. Moreover, there is a lack of theoretical guarantees for highly nonsmooth and nonconvex deep neural networks.

In recent years, to address these drawbacks, a recognized optimization scheme, namely the Alternating Direction Method of Multipliers (ADMM), has been introduced as an alternative to SGD for resolving deep learning problems. ADMM aims to divide a problem into several subproblems and then cooperate to achieve a global solution \cite{boyd2011distributed}. ADMM has been established as a successful approach to traditional machine learning  \cite{taylor2016training, tieleman2017divide,wang2020toward,wang2019admm,zeng2019global}. The following merits make ADMM a promising strategy: it carries out without gradient steps and circumvents gradient vanishing phenomena; it shows great scalability in the matter of the number of layers, training in parallel across the cores; it enables using the nondifferentiable activation functions and nonsmooth regularization term; it is resistant to the poor conditioning problems \cite{taylor2016training}. Thus, many researchers have explored the ADMM for various deep neural networks. It was primarily applied to resolve the Multi-Layer Perceptron (MLP) problem with convergence guarantees \cite{taylor2016training,wang2019admm}. Afterwards, it was expanded to other structures, including Recurrent Neural Networks (RNN) \cite{tang2020admmirnn}, and allowed parallel deep neural network training using multiple GPUs \cite{wang2020toward,guan2021pdladmm}.   

Despite the popularity of applying ADMM to deep neural networks, there exist some significant challenges: (\textit{i}) Cubic time complexity in respect of feature dimensions. This computational bottleneck stems from the matrix inversion used for updating the weight parameters. Additional sub-iterations are required for obtaining an inverse matrix that leads to the time complexity of $\mathrm{O}(n^3)$, in which $n$ is a feature dimension \cite{boyd2011distributed}. Taylor et al. \cite{taylor2016training} inferred from their experiments that more than 7000 cores are involved in training a neural network with 300 neurons based on ADMM. (\textit{ii}) Slow convergence towards a high accuracy solution. Even though ADMM is a competent optimizer for large-scale deep learning problems, it takes a long period to converge toward a solution with high precision. 
The convergence rate of deep learning via ADMM (dlADMM) methods \cite{wang2019admm} and block coordinate descent (BCD) algorithms in deep learning \cite{zeng2019global} is sublinear of $\mathrm{O}(1/k)$, where $k$ is the number of epochs. Recently, Wang et al. \cite{wang2022accelerated} improved the convergence rate by proposing an mDLAM algorithm, which can reach a linear convergence rate. However, to the best of the authors' knowledge, a theoretical framework to further speed up the convergence rate is not developed.

This study proposes a novel optimization structure to tackle these challenges: the Anderson Acceleration for Deep Learning Alternation Direction Method of Multipliers (AA-\linebreak DLADMM). To update the weight parameters, the quadratic estimation and backtracking algorithms are applied as an alternative to the matrix inversion, resulting in decreasing the time complexity from $\mathrm{O}(n^3)$ to $\mathrm{O}(n^2)$. Moreover, an acceleration method is used for nonconvex ADMM. The Anderson acceleration technique can accelerate the convergence rate by interpreting ADMM as a fixed-point iteration on an abstracted set of variables. Anderson acceleration exerts a quasi-Newton scheme, which seeks out an almost quadratic convergence rate. We evaluate our proposed AA-DLADMM algorithm on four benchmark datasets. Thanks to its acceleration approach and theoretical guarantee, our method attains better convergence than other comparison methods, including SGD and its variants and ADMM-based methods, in many experiments. Our key contributions in this paper are recapped as follows: 
\begin{itemize}
    \item We propose a novel optimization framework for deep neural networks. Backtracking and quadratic estimation are applied instead of matrix inversion.  
    \item We present an accelerated algorithm for deep neural networks based on nonconvex ADMM. The Anderson acceleration approach boosts the convergence rate to nearly quadratic. 
    \item We provide the convergence of our accelerated optimization algorithm subject to proper assumptions, surpassing the existing methods.
    \item We demonstrate the efficiency and performance of our proposed method by conducting comprehensive experiments on four benchmark datasets.
\end{itemize}

The rest of this paper is arranged as follows: In Section \ref{relatedwork}, we outline the recent research respecting this study. Section \ref{AA-DLADMM} describes the problem statement and presents the proposed AA-DLADMM algorithm for training deep neural networks. In Section \ref{convergence analysis}, we provide the details of the convergence properties of the proposed AA-DLADMM algorithm. In Section \ref{experiments}, we present the extensive experimental results to indicate and verify the efficiency of the proposed approach. Finally, Section \ref{conclusion} gives the conclusion of this work.

\section{Related Work} \label{relatedwork}
The literature on deep learning optimization techniques is divided into two main categories: SGD algorithms and alternating minimization schemes, which are outlined as follows:

\textbf{SGD algorithms:} Rumelhart et al. \cite{rumelhart1986learning} are the first researchers to introduce the well-known backpropagation algorithm to effectually train neural networks. In 2012, deep learning attracted the attention of many researchers owing to the high accuracy of AlexNet \cite{krizhevsky2017imagenet}. Consequently, various GD-based optimization algorithms have been presented to accelerate the convergence rate, such as Polyak momentum \cite{polyak1964some}, initialization, and Nesterov momentum  \cite{sutskever2013importance}. In the past decades, plenty of SGD methods that integrated adaptive learning rates have been proposed, in particular, AdaGrad \cite{duchi2011adaptive}, RMSProp \cite{tieleman2017divide}, Adam \cite{kingma2014adam}, AMSGrad \cite{reddi2019convergence}, Adabelief \cite{zhuang2020adabelief}, and Adabound \cite{luo2019adaptive}. The convergence assumption of these adaptive optimizers does not apply to deep learning problems due to the existence of the ReLU as a non-differentiable activation function.

\textbf{Alternating minimization:} Many recent studies have exerted the Alternating Direction Method of Multipliers \linebreak (ADMM) \cite{boyd2011distributed}, auxiliary coordinate (MAC) \cite{carreira2014distributed}, and Block Coordinate Descent (BCD) \cite{zeng2019global} for constrained neural network problems. In these algorithms, the gradient vanishing problem has not occurred, and non-differentiable activation functions, e.g., ReLU and nonsmooth regularization terms, can be adopted. 
Li et al. \cite{li2021community} introduced an ADMM-based algorithm to acquire distributed learning for Graph Convolutional Network (GCN).

\textbf{ADMM for nonconvex problem:} Despite the leading performance of ADMM for nonconvex problems, its convergence property is not profound, caused by the complexity of equality and inequality constraints and coupled objectives. Wang et al. \cite{wang2019global} considered mild convergence assumptions for coupled and nonsmooth objective functions of nonconvex ADMM. Chen et al. \cite{chen2019extended} explored the application of ADMM for quadratic coupling problems. Wang et al. \cite{wang2017nonconvex} analyzed the convergence of nonconvex ADMM in the presence of constrained equality problems. Although ADMM has been implemented in various deep learning problems, a lack of theoretical convergence analysis has remained. 

\textbf{Anderson acceleration:} Anderson acceleration \cite{walker2011anderson} is an acknowledged approach for increasing the speed of convergence of a fixed-point iteration. It has a wide variety of applications in computational physics \cite{lipnikov2013anderson}, numerical linear algebra \cite{pratapa2016anderson}, and robotics \cite{pavlov2018aa}. The main conception of Anderson acceleration is using $m$ previous iterations to estimate a new iterate so that it converges more rapidly to the fixed point. It should be noted that this method is a quasi-Newton technique to retrieve the root of the residual function \cite{eyert1996comparative}. Recently, Zhang et al. \cite{zhang2019accelerating} accelerated the convergence of nonconvex ADMM by applying Anderson acceleration in computer graphics.

\section{The AA-DLADMM Algorithm} \label{AA-DLADMM}

In this section, we present our proposed AA-DLADMM algorithm. \hyperlink{thesentence}{Table \ref{notation}} describes the notation utilized in this study. Section \ref{problemform} provides an overview of the existing ADMM-based method for deep learning and then discusses the AA-DLADMM algorithm in detail by reformulating the problem. Section \ref{algorithm update} investigates the procedure of updating parameters.

\begin{table}[h!]
\begin{center}
\caption{Notations utilized in this paper}
\label{notation}
\begin{tabular}{p{1cm} p{6.8cm} }
\toprule
Notations & Descriptions\\                    
\midrule
$L$         & Number of layers                   \\ 
$W_l$          & The weight matrix in the $l^{th}$ layer        \\ 
$b_l$        & The intercept vector in the $l^{th}$ layer        \\
$z_l$         & The output of linear mapping for the $l^{th}$ layer  \\ 
$h_l(z_l)$         & The nonlinear activation function for the $l^{th}$ layer                 \\ 
$a_l$         & The output of the $l^{th}$ layer                 \\ 
$x$         & The input matrix of neural networks                 \\ 
$y$         & The predefined label vector                 \\ 
$R(z_l;y)$         & The loss function for the $l^{th}$ layer                 \\ 
$\Omega_l(W_l)$         & The regularization term for the $l^{th}$ layer                 \\ 
$\epsilon$         & The tolerance for the nonlinear mapping                 \\ 
$n_l$         & The number of neurons in the $l^{th}$ layer                \\ 
\midrule
\end{tabular}
\end{center}
\end{table}

\subsection{Problem formulation} \label{problemform}

Generally, MLP includes $L$ layers in which linear mapping and nonlinear activation functions are used. A linear mapping consists of a weight vector $W_l \in \mathbb{R}^{n_l \times n_{l-1}}$, and a nonlinear mapping is determined by a continuous activation function $h_l(\bullet)$. The outputs of the $l$-th layer is $a_l=h_l(W_l a_{l-1})$ with the input of the $(l-1)$-th layer as $a_{l-1} \in \mathbb{R}^{n_{l-1}}$. The neural network problem is mathematically formulated in Problem 1 below by inserting the auxiliary variable $z_l$ as the output of the linear mapping:

\textbf{Problem 1.}
\begin{equation}
\begin{split}
    & \min_{\substack{ W_l, z_l, a_l, b_l}} R(z_L;y)+\sum_{l=1}^L \Omega_l(W_l)\\
    & s.t. \hspace{0.5cm} z_l=W_l a_{l-1}+b_l \hspace{0.2cm} (l=1, \ldots, L),\\
    & \hspace{1cm}a_l=h_l(z_l) \hspace{0.35cm} (l=1, \ldots, L-1).
\end{split}
\end{equation}

where $a_0 =x \in \mathbb{R}^{d}$ is the input of the neural network with $d$ as the number of feature dimensions, and $y$ denotes the presumed label vector. The function $R(z_L;y)$ represents the loss function in the $L$-th layer, and $\Omega_l(W_l) \geq 0$ is a regularization term in the $l$-th layer. Both $R$ and $\Omega_l$ functions are continuous, convex, and proper.

In Problem 1, the nonconvex equality constraint $a_l=h_l(z_l)$ is the most demanding part since the nonlinear activation function is used. Therefore, solving the $z_l$ subproblem and reaching an optimal solution is challenging. Furthermore, there is a lack of guarantee for resolving the nonlinear equality constraints according to alternating minimization optimizers. To overcome the above-mentioned challenges, $h_l(z_l)$ is assumed to be quasilinear \cite{wang2022accelerated}. Thus, nonlinear activation functions such as smooth sigmoid, tanh, and the Rectified Linear Unit (ReLU) can be utilized. Accordingly, Problem 1 can be reformulated with inequality constraints by introducing a tolerance $\epsilon$ as follows:

\textbf{Problem 2.}
\begin{equation}
\begin{split}
    & \min_{\substack{ W_l, z_l, a_l,b_l}}  R(z_L;y)+\sum_{l=1}^L \Omega_l(W_l)\\
    & s.t. \hspace{0.5cm} z_l=W_l a_{l-1}+b_l \hspace{0.2cm} (l=1,\cdots, L),\\
    & \hspace{1cm}  h_l(z_l)-\epsilon\leq a_l \leq h_l(z_l)+\epsilon \hspace{0.2cm} (l=1,\cdots, L-1).
\end{split}
\end{equation}

where $\textbf{W}=\{W_l\}_{l=1}^L$, $\textbf{z}=\{z_l\}_{l=1}^L$, $\textbf{a}=\{a_l\}_{l=1}^{L-1}$, $\textbf{b}=\{b_l\}_{l=1}^L$, and  $\epsilon$ is exerted to project the nonconvex constraints into $\epsilon$-balls. In Problem 2, the linear constraint $z_l=W_l a_{l-1}+b_l$ can be relaxed with an $\ell_2$ penalty term in the objective function, which is formulated as follows:

\textbf{Problem 3.}
\begin{equation}
\begin{split}
    & \min_{\substack{W_l, z_l, a_l, b_l}} \mathcal{F}(\textbf{W}, \textbf{z}, \textbf{a}, \textbf{b})= R(z_L;y)+\sum_{l=1}^L \Omega_l(W_l)\\
    & \hspace{1.3cm} + \frac{\rho}{2} \sum_{l=1}^L \norm{z_l- W_l a_{l-1}-b_l}_2^2\\
    & s.t.  \hspace{0.3cm}  h_l(z_l)-\epsilon\leq a_l \leq h_l(z_l)+\epsilon \hspace{0.3cm} (l=1,\cdots, L-1).
\end{split}
\end{equation}

where $\rho>0$ is a penalty parameter and as $\rho \rightarrow \infty$ and $\epsilon \rightarrow \infty$, Problem 3 approaches Problem 1. Indeed, Problem 3 is still nonconvex due to the existence of $h_l(z_l)$, but it is convex in terms of one variable while others are fixed. To give an example, Problem 3 is convex regarding $\{z\}_{l=1}^L$ when $\{W\}_{l=1}^L$ and $\{a\}_{l=1}^L$ set to be fixed \cite{wang2022accelerated}.

There is an implicit issue in solving Problem 3, which relates to having a slow convergence rate. The convergence rate of the dlADMM and mDLAM algorithms is sublinear to $\mathrm{O}(1/k)$ and linear, respectively, to solve Problem 3 \cite{wang2019admm,wang2022accelerated}. To tackle this issue, we apply the Anderson acceleration methods to elevate the convergence rate.

\subsection{Anderson Acceleration}

The principle of devising a quadratically convergent method that applies solely first-order derivatives is derived from Nesterov’s technique to address convex optimization problems. In this method, linear extrapolation of the two previous iterations is assessed for executing iterations instead of the current result exclusively \cite{nesterov27method}. Even though such a momentum approach can be utilized to speed up the ADMM method, it is improper for nonconvex problems due to the extrapolated point's procedure, which can not be governed by a bound on the objective function's curve. We summarize the Anderson acceleration below and for more information, you can refer to \cite{goldstein2014fast}.

Consider a fixed-point iteration
\begin{equation} \label{fixed}
    x_{k+1}=G(x_k)
\end{equation}
where $x \in \mathbb{R}^n$ and $G: \mathbb{R}^n \rightarrow \mathbb{R}^n$. The purpose of Anderson acceleration is to achieve an almost quadratic convergence rate by deducting the quasi-Newton strategy. For a solution $x^*$ to the fixed-point iteration (\ref{fixed}), the residual
\begin{equation} \label{residual}
    F(x)=G(x)-x
\end{equation}
must become zero. The fundamental idea of Anderson acceleration is to utilize the residuals of the recent iterate $x_k$ and its previous $m$ iterates $x_{k-1}, ..., x_{k-m}$ to extract a new iterate $x_{k+1}^{AA}$, which diminish the residual to the furthest extent. Particularly, an affine subspace can be spanned by $x_k, x_{k-1},\cdots, x_{k-m}$, where each point can be defined as
\begin{equation}
    x(\alpha)=x_k + \sum_{j=1}^m \alpha_j (x_{k-j}-x_k)
\end{equation}
and $\alpha=(\alpha_1, \cdots, \alpha_m)$ denotes affine coordinates. The new step is obtained by this subspace under the fixed-point mapping $G$:
\begin{equation}\label{xk+1}
    x_{k+1}^{AA}=G(x_k)- \sum_{j=1}^m \alpha_j^* (G(x_{k-j+1}) -G(x_{k-j}))
\end{equation}
where the coefficients $\alpha_j^*$, $j=1, \ldots, m$ are acquired by solving a linear least-squares problem:
\begin{equation} \label{leastsquare} \begin{split}
    (\alpha_1^*, \cdots , \alpha_m^*)= \operatorname*{argmin}_{\alpha_1, ..., \alpha_m} &\Biggl|\!\Biggl| F(x_k)- \sum_{j=1}^m \alpha_j \Bigl(F(x_{k-j+1}) \\& - F(x_{k-j})\Bigr)\Biggr|\!\Biggr|^2
    \end{split}
\end{equation}
Here $F(x_k)=G(x_k)-x_k$ represents the residual at step $k$. We define $\Delta F_k=F_{k+1}-F_{k}$ and $\Delta G_{k-j}=G(x_{k-j+1})-G(x_{k-j})$, where $F_k=F(x_k)$. The least-squares problem (\ref{leastsquare}) can be solved as follows concerning the linear dependency of $\{\Delta F_{k-j}\}$ \cite{fang2009two}: 
\begin{equation} \label{normal}
    (D^\top D) \alpha=D^\top F_k
\end{equation}
where $D=[\Delta F_{k-1}, \cdots, \Delta F_{k-m}]$.

The relationship between the Anderson acceleration and ADMM is that the ADMM algorithm can be noticed as fixed-point iterations $(x_k,y_k) \rightarrow (x_{k+1}, y_{k+1}), k=0,1,2, \cdots$. To guarantee the stability of Anderson acceleration, some metrics should be determined to assess the efficiency of an accelerated iteration and a fall-back plan since the nonconvex constraints in the optimization are not met. Goldstein et al. \cite{goldstein2014fast} indicated that if the problem is convex, the integrated residual is monotonously reduced via ADMM.

\subsubsection{Choice of parameter $m$}

Fang et al. \cite{fang2009two} asserted that Anderson acceleration can be assumed as a quasi-Newton technique to obtain the root of the residual function, using the $m$ previous iterates to estimate the inverse Jacobian. Thus, the choice of parameter $m$ has an impact on the performance of acceleration. We noticed that a larger value of $m$ contributes to using more information for estimating the inverse Jacobian, which decreases the required number of epochs for convergence. However, a larger $m$ soars the computational cost, and then the training procedure is influenced by overfitting, which is out of range. To this end, we experimentally choose $m=8$ for our experiments. 

\subsubsection{Analysis of Anderson acceleration}
This subsection summarizes the Anderson acceleration algorithm \cite{tang2022fast}.
The following matrix of secants in $x$ is built based on the outcomes of the previous $m$ iterations recollected from memory in each epoch $k$.
\begin{equation} \label{matrix secants}
\begin{split}
    & \xi_i^x= x_{i+1} -x_i\\
    &\Xi_k^x = [\xi_{k-m_k}^x \cdots \xi_{k-1}^x]\\
    & \xi_i^F= F(x_{i+1}) -F(x_i)\\
    & \Xi_k^F=[\xi_{k-m_k}^F \cdots \xi_{k-1}^F]
\end{split}
\end{equation}
where $m_k$ denotes the memory in iteration $k$. As it is stated in \cite{fang2009two}, the estimated Jacobian of $F$ minimizing $\norm{B_k - I}_F$ subject to the condition $B_k \Xi_k^x=\Xi_k^F$ can be written as
\begin{gather}\label{Bk}
    B_k= I +(\Xi_k^F - \Xi_k^x)(\Xi_k^{x\top} \Xi_k^x)^{-1} \Xi_k^{x\top} \\ 
    B_k^{-1}= I +(\Xi_k^x - \Xi_k^F)(\Xi_k^{x\top} \Xi_k^F)^{-1} \Xi_k^{x\top}
\end{gather}
Thus, the quasi-Newton iteration is defined as 
\begin{equation}
    x_{k+1}=x_{k}-B_k^{-1}F_k
\end{equation}

However, if either $\Xi_k^x$ or $\Xi_k^F$ suffers from rank deficiency, the iteration may be unable to carry out. To deal with this issue, \cite{zhang2020globally} fixed $B_k$ using its limitation to span ($\Xi_k^x$) and its orthogonal complement. Thus, in Eq. (\ref{Bk}) $B_k$ can be built inductively beginning from $B_k^0 =I$ as
\begin{equation}\label{bk ortho}
    B_k^{i+1}=B_k^i + \frac{(\xi_{k-m_k +i}^F -B_k^i \xi_{k-m_k +i}^x)\hat{\xi}_{k-m_k+i}^{x\top}}{\hat{\xi}_{k-m_k+i}^{x\top} \xi_{k-m_k +i}^x}
\end{equation}
where $i=0, \cdots, m_k-1$ and $B_k=B_k^{m_k}$. In addition, $\{\hat{\xi}_i^x\}_{i=k-m_k}^{k-1}$ can be acquired from $\{\xi_i^x\}_{i=k-m_k}^{k-1}$ via applying Gram-Schmidt orthogonalization:
\begin{equation}
    \hat{\xi}_i^x=\xi_i^x - \sum_{j=k-m_k}^{i-1} \frac{\hat{\xi}_j^{x\top} \xi_i^x}{\hat{\xi}_j^{x\top} \xi_j^x} \hat{\xi}_j^x, \hspace{0.3cm} i=k-m_k, \cdots, k-1.
\end{equation}
Note that (\ref{bk ortho}) becomes singular when $\hat{\xi}_{k-m_k +i}^x =0$, while $\xi_{k-m_k +i}^x \neq 0$. In fact, it is possible even if $m_k$ is selected concerning $\min \{m,k\}$ for $1 \leq m \leq \infty$, which $m$ represents the full memory. To tackle this problem, a restart checking procedure is utilized to free the memory instantly before the algorithm is adjacent to the slump. To be more distinct, $m_k$ is increasing up to either $m_k=m+1$ for some integer $1 \leq m \leq \infty$, or the Gram-Schmidt orthogonalization will be ill-conditioned $\norm{\hat{\xi}_{k-1}^x}_2 < \tau \norm{\xi_{k-1}^x}_2$ for $\tau \in (0,1)$, in which circumstance $m_k$ is reset to zero. Furthermore, to avoid the possible singularity of $B_k$, \cite{zhang2020globally} defined $\overline{\theta} \in (0,1)$ to replace $\xi_{k-m_k +i}^F$ in (\ref{bk ortho}) with
\begin{equation} \label{new xi}
    \Tilde{\xi}_{k-m_k +i}^F = \theta_k^i \xi_{k-m_k +i}^F +(1-\theta_k^i) B_k^i \xi_{k-m_k+i}^x
\end{equation} 
where $\theta_k^i= \psi_{\overline{\theta}} \left(\frac{\hat{\xi}_{k-m_k+i}^{x\top} (B_k^{-1}) \xi_{k-m_k+i}^F}{\norm{\hat{\xi}_{k-m_k+i}^{x\top}}_2^2}\right)$, and $\psi_{\overline{\theta}}(\eta)$ is determined by
\begin{equation}
    \psi_{\overline{\theta}}(\eta)= \begin{cases}
        (1-\mbox{sign}(\eta) \overline{\theta})/(1-\eta)  \hspace{0.5cm} \mathrm{if} \hspace{0.1cm} |\eta|< \overline{\theta}\\
        1 \hspace{3.5cm} \mathrm{if} \hspace{0.1cm} |\eta|\geq \overline{\theta}
    \end{cases}
\end{equation}

Notice that we have $B_k^i=B_{k-m_k+i}$ as restart checking is employed to clear the memory and $B_k$ is computed in (\ref{bk ortho}). Furthermore, we can eliminate the requirement for keeping updates for $B_k^i$ in the regularization by observing that $B_k^i \xi_{k-m_k+i}^x =B_{k-m_k+i} \xi_{k-m_k+i}^x = -B_{k-m_k+i} B_{k-m_k+i}^{-1} F_{k-m_k+i}= -F_{k-m_k+i}$.
\begin{lemma} \label{lemma1}
    (Anderson acceleration's well-conditioned \cite[Lemma 3]{zhang2020globally}) By choosing the $m_k$ from the restart checking strategy and considering (\ref{new xi}), we obtain $\norm{B_k} \leq 3(1+ \overline{\theta}+\tau )^{m} /\tau^m -2$ for all $k \geq 0$. 
\end{lemma}
Therefore, the rank deficiency issue is thoroughly resolved. Additionally, the inverse of $B_k$ remains bounded as we have 
\begin{equation} \label{Bk-1 norm}
    \norm{B_k^{-1}}_2 \leq \frac{ \left( 3 \left(\frac{1+\overline{\theta}+\tau}{\tau} \right)^m -2 \right)^{n-1}} {\overline{\theta}^m}.
\end{equation}
where $n$ is the dimension of $x$.

Moreover, to guarantee the decline of the residual norm, $F$, we apply a safeguarding strategy developed by \cite{zhang2020globally} to investigate that the residual norm is adequately small.

Finally, the Anderson acceleration approach for the proposed AA-DLADMM is summarized in Algorithm 1. Lines 4-7 implement the restart checking scheme, lines 8-11 execute the regularization, lines 12-13 perform the safeguarding method, and $r_{k+1}=z_{k+1}-W_{k+1} a_k -b_{k+1}$.
Please notice that the Anderson acceleration recommends an update $\Tilde{x}_k$ to ensure that $B_{k-1} \xi_{k-1}^x =-B_{k-1} B_{k-1}^{-1} F_{k-1}=-F_{k-1}$ is still retained, resulting in the maintenance of updating $B_k$ being discarded. As a result, in line 4 of Algorithm 1, we applied $\Tilde{x}_k$ instead of defining $\xi_{k-1}^x= x_k - x_{k-1}$ and $\xi_{k-1}^F= F(x_k)- F(x_{k-1})$.

\subsection{Algorithm Updates} \label{algorithm update}

The proposed AA-DLADMM algorithm contains the following subproblems, which will be introduced in detail. The quadratic approximation and backtracking methods are employed to solve the subproblems to avoid matrix inversion.

     \textbf{1. Update $W_l$} \\  The variables $W_l (l=1, \cdots, L)$ are updated as follows:
    \begin{equation} \label{update w}
        W_l^{k+1} \leftarrow \operatorname*{argmin}_{W_l} \phi\left(a_{l-1}^{k+1}, W_l, z_l^k,b_l^k\right) +\Omega_l(W_l).
    \end{equation}
    where $\phi(a_{l-1}^{k+1}, W_l, z_l^k,b_l^k)=\frac{\rho}{2} \norm{z_l- W_l a_{l-1}- b_l}_2^2$. Due to the coupling of $W_l$ and $a_{l-1}$ in $\phi(\bullet)$, the inversion execution of $a_{l-1}^{k+1}$ is required for solving $W_l$, which is computationally costly. To overcome this challenge, similarly to the dlADMM algorithm \cite{wang2019admm}, we define $\mathcal{P}_l(W_l; \theta_l^{k+1})$ as a quadratic estimation of $\phi$ at $W_l^k$, which can be written as follows:
    \begin{equation}
    \begin{split}
       \mathcal{P}_l^{k+1} \left(W_l;\theta_l^{k+1}\right) &= \phi \left(a_{l-1}^k, \overline{W}_l^{k+1}, z_l^k,b_l^k \right) +(\nabla_{\overline{W}_l^{k+1}} \phi)^\top\\
       &\left(W_l - \overline{W}_l^{k+1}\right) + \frac{\theta_l^{k+1}}{2} \norm{W_l -\overline{W}_l^{k+1}}_2^2
    \end{split}
    \end{equation}
where $\overline{W}_l^{k+1}=W_l^k$ and $\theta_l^{k+1} >0$ denotes a scalar parameter, which is adopted by the backtracking algorithm \cite{wang2019admm} to satisfy the condition below
\begin{equation} \label{p}
    \mathcal{P}_l^{k+1} \left(W_l;\theta_l^{k+1}\right) \geq \phi\left(a_{l-1}^k, \overline{W}_l^{k+1}, z_l^k, b_l^k \right),     
\end{equation}
and $\nabla_{\overline{W}_l^{k+1}} \phi$ is the gradient of $W_l$ at $\overline{W}_l^{k+1}$. Thus, we minimize the following rather than minimizing (\ref{update w}).
\begin{equation} \label{quadratic w}
     W_l^{k+1} \leftarrow \operatorname*{argmin}_{W_l} \mathcal{P}_l^{k+1} \left(W_l;\theta_l^{k+1}\right) +\Omega_l(W_l),
\end{equation}
Eq. (\ref{quadratic w}) holds a closed-form solution for assuming either $\textit{l}_1$ or $\textit{l}_2$ regularization.

\begin{table}[t]
\begin{center}
\label{anderson}
\begin{tabular}{p{8cm} }
\toprule
\textbf{Algorithm 1:} Anderson acceleration algorithm \\                    
\midrule
1: Initialization: $x_{\mathrm{default}}=x_0$ ; $r_{\mathrm{prev}}=\infty$; $k=0$; reset=TRUE; safeguarding constants $d$, $\varepsilon>0$; $\alpha, \overline{\theta}, \tau \in (0,1)$; $m_0=n_{AA}=0$; $\bar{U}=\norm{F(x_0)}_2$ ; $x_1=\Tilde{x}_1 = (1-\alpha) x + \alpha G(x)$  \\ 
2: \textbf{if} reset==TRUE \textbf{OR} $r < r_\mathrm{prev}$ \textbf{then}   \\
 \hspace{0.2cm} // Record the recent accepted iterate\\
3:  \hspace{0.2cm} $x_k=x_{\mathrm{default}}$; $r_{\mathrm{prev}}=r$; reset=FALSE\\
 \hspace{0.2cm} // Compute the accelerated iterate\\
4:  \hspace{0.2cm} $\xi_{k-1}^x=\tilde{x}_{k}-x_{k-1}$; $\xi_{k-1}^F=F(\tilde{x}_{k})-F(x_{k-1})$.\\
5: \hspace{0.2cm} $\hat{\xi}_{k-1}^x=\xi_{k-1}^x - \sum_{j=k-m_{k}}^{k-2} \frac{\hat{\xi}_j^{x\top} \xi_{k-1}^x}{\hat{\xi}_j^{x\top} \hat{\xi_j}^x} \hat{\xi_j^x}$ \\ 
6: \hspace{0.2cm} if $m_k=m+1$ or $\norm{\hat{\xi}_{k-1}^x}_2 < \tau \norm{\xi_{k-1}^x}_2$  \\
7: \hspace{0.9cm} reset $m_k=0$, $\hat{\xi}_{k-1}^x=\xi_{k-1}^x$, and $B_{k-1}^{-1}=I$. \\
8: \hspace{0.2cm}  Compute $\bar{\xi}_{k-1}^F=\theta_{k-1} \xi_{k-1}^F -(1-\theta_{k-1}) F_{k-1}$ \\
9: \hspace{0.25cm} with $\theta_{k-1}=\psi_{\bar{\theta}}(\eta_{k-1})$, and $\eta_{k-1}=\hat{\xi}_{k-1}^{x\top} B_{k-1}^{-1} \xi_{k-1}^F/\norm{\hat{\xi}_{k-1}^x}^2$. \\
10:\hspace{0.2cm} Update $B_k^{-1}=B_{k-1}^{-1} + \frac{(\xi_{k-1}^x-B_{k-1}^{-1} \bar{\xi}_{k-1}^F)\hat{\xi}_{k-1}^{x\top} B_{k-1}^{-1}}{\hat{\xi}_{k-1}^{x\top} B_{k-1}^{-1} \bar{\xi}_{k-1}^F}$\\
11:\hspace{0.8cm}, and $\tilde{x}_{k+1}= x_k - B_k^{-1} F(x_k)$.  \\
12:\hspace{0.2cm} if $\norm{F(x_k)} \leq d\bar{U}(n_{AA}+1)^{-(1+\varepsilon)}$ \\
13:\hspace{0.9cm} $x_{k+1}=\tilde{x}_{k+1}, n_{AA}=n_{AA}+1$ \\ 
14:\textbf{else}\\
\hspace{0.2cm} // Revert to the last accepted iterate \\
15:\hspace{0.2cm} reset=TRUE, $x_k =x_{\mathrm{default}}$\\

\midrule
\end{tabular}
\end{center}
\end{table}

\textbf{2. Update $a_l$}

The variables $a_l(l=1, \cdots, L-1)$ are updated as follows:
\begin{equation}\begin{split}\label{update a}
    a_l^{k+1} &\leftarrow  \operatorname*{argmin}_{a_l} \phi \left(a_{1}, W_{l+1}^k, z_{l+1}^k, b_l^{k+1} \right) \\
    & s.t. \hspace{0.3cm} h_l(z_l^{k+1}) -\epsilon \leq a_l \leq h_l(z_l^{k+1}) +\epsilon
\end{split} 
\end{equation}
Likewise, we introduce $\mathcal{Q}_l^{k+1}(a_l;\tau_l^{k+1})$ as a quadratic approximation of $\phi$ at $a_l^{k+1}$. It can be written mathematically as follow:
\begin{equation}
    \begin{split}
        \mathcal{Q}_l^{k+1} \left(a_l;\tau_l^{k+1} \right)&= \phi \left(a_l^{k+1}, W_{l+1}^k,z_{l+1}^k, b_{l+1}^k\right)+ (\nabla_{a_l^{k+1}} \phi)^\top \\
        & \left(a_l -a_l^{k+1}\right) + \frac{\tau_l^{k+1}}{2} \norm{a_l -a_l^{k+1}}_2^2
    \end{split}
\end{equation}
where $\tau_l^{k+1}$ is a scalar parameter selected by the backtracking strategy \cite{wang2019admm} to satisfy the following condition
\begin{equation}
    \mathcal{Q}_l^{k+1} \left(a_l;\tau_l^{k+1} \right) \geq \phi \left(a_l^{k+1}, W_{l+1}^k,z_{l+1}^k, b_l^{k+1}\right).
\end{equation}
Therefore, we minimize the following problem instead of solving the minimization problem in (\ref{update a})
\begin{equation}
    \begin{split}\label{quadratic a}
    a_l^{k+1} &\leftarrow  \operatorname*{argmin}_{a_l}  \mathcal{Q}_l^{k+1}\left(a_l;\tau_l^{k+1}\right) \\
    & s.t. \hspace{0.3cm} h_l(z_l^{k+1}) -\epsilon \leq a_l \leq h_l(z_l^{k+1}) +\epsilon
\end{split} 
\end{equation}

The solution to (\ref{quadratic a}) is acquired as follows by assuming $L^{k+1}$ and $U_l^{k+1}$ to be the lower and upper bound of the constraint $ \{h_l(z_l^{k+1}) -\epsilon \leq a_l \leq h_l(z_l^{k+1}) +\epsilon \}$, respectively.
\begin{equation}
    a_l^{k+1} \leftarrow \min \Big(\max \big(L^{k+1} , a_l^{k+1} - \nabla_{a_l^{k+1}} \phi / \tau_l^{k+1}\big), U_l^{k+1}\Big).
\end{equation}

\textbf{3. Update $z_l$}

The variables $z_l(l=1, \cdots, L-1)$ are updated as follows:
\begin{equation}\begin{split}\label{update z}
    &z_l^{k+1} \leftarrow  \operatorname*{argmin}_{z_l} \phi\left(a_{1-1}^k, W_{l}^{k+1}, z_{l}, b_l^{k+1}\right) \\
    & \hspace{0.8cm} s.t. \hspace{0.3cm} h_l(z_l^{k+1}) -\epsilon \leq a_l \leq h_l(z_l^{k+1}) +\epsilon,
\end{split} 
\end{equation}
and for $z_L^{k+1}$
\begin{equation} \label{updatzL}
    z_L^{k+1} \leftarrow \operatorname*{argmin}_{z_L} \phi\left(a_{L-1}^{k+1}, W_{L}^{k+1}, z_{L}, b_L^{k+1}\right)+ R(z_L;y).
\end{equation}
Similarly, we define $\mathcal{V}_l^{k+1}(z_l)$ as follows:
\begin{equation}
    \begin{split}
        \mathcal{V}_l^{k+1}(z_l)&= \phi \left(a_{l-1}^{k+1}, W_{l}^{k+1},z_{l}^{k+1}, b_l^{k+1}\right)+ (\nabla_{z_l^{k+1}} \phi)^\top \\
        & \left(z_l -z_l^{k+1}\right) + \frac{\rho}{2} \norm{z_l -z_l^{k+1}}_2^2
    \end{split}
\end{equation}
where $\rho$ is the penalty parameter. Thus, the following problems should be solved instead of (\ref{update z}) and (\ref{updatzL}).
\begin{equation}\begin{split}\label{quadratic z}
    &z_l^{k+1} \leftarrow  \operatorname*{argmin}_{z_l} \mathcal{V}_l^{k+1}(z_l) \\
    & \hspace{0.8cm} s.t. \hspace{0.3cm} h_l(z_l^{k+1}) -\epsilon \leq a_l \leq h_l(z_l^{k+1}) +\epsilon \hspace{0.2cm} (l<L),
\end{split} 
\end{equation}
\begin{equation} \label{quadratizL}
    z_L^{k+1} \leftarrow \operatorname*{argmin}_{z_L} \mathcal{V}_L^{k+1}(z_L)+ R(z_L;y).
\end{equation}
The solution for $z_l(l=1, \cdots, L-1)$ is attained by
\begin{equation}
    z_l^{k+1} \leftarrow \min \Big(\max \big(L^{k+1} , z_l^{k+1} - \nabla_{z_l^{k}} \phi / \rho\big), U_l^{k+1}\Big).
\end{equation}
On account of the convexity of $\mathcal{V}(\bullet)$ and $R(\bullet)$ with respect to $z_L$, (\ref{quadratizL}) is a convex problem. Accordingly, the Fast Iterative Soft-Thresholding Algorithm (FISTA) \cite{beck2009fast} can be used to solve (\ref{quadratizL}).  

\textbf{4. Update $b_l$}

The variables $b_l(l=1, \cdots, L)$ are updated as follows:
\begin{equation}\label{update b}
    b_l^{k+1} \leftarrow  \operatorname*{argmin}_{b_l} \phi\left(a_{1-1}^{k+1}, W_{l}^{k+1}, z_{l}^{k+1}, b_l \right).      
\end{equation}
Similarly to update $z_l^{k+1}$, we define $\mathcal{U}_l^{k+1}(b_l; \rho)$ as a quadratic approximation of $\phi(\bullet)$ at $b_l^k$, which can be mathematically reformulated as follows:
\begin{equation}
    \begin{split}
        \mathcal{U}_l^{k+1}(b_l; \rho)&= \phi \left(a_{l-1}^{k+1}, W_{l}^{k+1},z_{l}^{k+1}, b_l^{k+1}\right)+ (\nabla_{b_l^{k+1}} \phi)^\top \\
        & \left(b_l -b_l^{k+1}\right) + \frac{\rho}{2} \norm{b_l -b_l^{k+1}}_2^2.
    \end{split}
\end{equation}
Hence, the following problem is minimized instead of minimizing (\ref{update b}).
\begin{equation} \label{quadratibL}
    b_l^{k+1} \leftarrow \operatorname*{argmin}_{b_l} \mathcal{U}_l^{k+1}(b_l;\rho).
\end{equation}
The solution to (\ref{quadratibL}) can be obtained as follows since it is convex and has a closed-form solution.
\begin{equation}
    b_l^{k+1} \leftarrow b_l^{k+1} -\nabla_{b_l^{k+1}} \phi / \rho
\end{equation}









\section{Convergence Analysis} \label{convergence analysis}

In this section, we investigate the convergence of the proposed AA-DLADMM algorithm. The following assumptions are deemed for the convergence analysis of the AA-DLADMM algorithm.
\begin{assumption} \label{assumption1}
    $\mathcal{F}(\textbf{W}, \textbf{z}, \textbf{a}, \textbf{b})$ is coercive over the domain 
    $$\mathcal{D}=\{(\textbf{W}, \textbf{z}, \textbf{a}, \textbf{b})| h_l(z_l)-\epsilon \leq a-l \leq h_l(z_l) + \epsilon,~ l=1, \cdots, L-1)\}.$$
    To put it simply, if $(\textbf{W}, \textbf{z}, \textbf{a}, \textbf{b}) \in \mathcal{D}$ and $\norm{(\textbf{W}, \textbf{z}, \textbf{a}, \textbf{b})} \rightarrow \infty$, then we have $\mathcal{F}(\textbf{W}, \textbf{z}, \textbf{a}, \textbf{b}) \rightarrow \infty$. Furthermore, $R(z_l;y)$ is Lipschitz differentiable with Lipschitz constant $\mathcal{H} \geq 0$.
\end{assumption}


Assumption~\ref{assumption1} is a mild assumption as common loss functions, including the least square loss and the cross-entropy loss, satisfy this Assumption. 

\begin{lemma} \label{lemma2} There exist $\beta_l, \delta_l, \gamma_l,\zeta_l>0$ for $k\in \mathbb{N}$ such that it holds
\begin{equation} \label{f(w)}
\begin{split}
    \mathcal{F}\big(\textbf{W}_{l-1}^{k+1}, \textbf{z}_{l-1}^{k+1}, \textbf{a}_{l-1}^{k+1}&, \textbf{b}_{l-1}^{k+1}\big) - \mathcal{F}\big(\textbf{W}_{l}^{k+1}, \textbf{z}_{l-1}^{k+1}, \textbf{a}_{l-1}^{k+1}, \textbf{b}_{l-1}^{k+1}\big)\\& \geq \frac{\beta_l^{k+1}}{2} \norm{W_{l}^{k+1} - W_l^k}_2^2.\\
        \end{split}
\end{equation}
\begin{equation} \label{f(z)}
\begin{split}
    \mathcal{F}\big(\textbf{W}_{l}^{k+1}, \textbf{z}_{l-1}^{k+1}, \textbf{a}_{l-1}^{k+1}&, \textbf{b}_{l-1}^{k+1}\big) - \mathcal{F}\big(\textbf{W}_{l}^{k+1}, \textbf{z}_{l}^{k+1}, \textbf{a}_{l-1}^{k+1}, \textbf{b}_{l-1}^{k+1}\big)\\ &\geq \frac{\delta_l^{k+1}}{2} \norm{z_{l}^{k+1} - z_l^k}_2^2.\\
        \end{split}
\end{equation}
\begin{equation}\label{f(a)}
\begin{split}
    \mathcal{F}\big(\textbf{W}_{l}^{k+1}, \textbf{z}_{l}^{k+1}, \textbf{a}_{l-1}^{k+1}&, \textbf{b}_{l-1}^{k+1}\big) - \mathcal{F}\big(\textbf{W}_{l}^{k+1}, \textbf{z}_{l}^{k+1}, \textbf{a}_{l}^{k+1}, \textbf{b}_{l-1}^{k+1}\big)\\& \geq \frac{\gamma_l^{k+1}}{2} \norm{a_{l}^{k+1} - a_l^k}_2^2.\\
        \end{split}
\end{equation}
\begin{equation} \label{f(b)}
\begin{split}
    \mathcal{F}\big(\textbf{W}_{l}^{k+1}, \textbf{z}_{l-1}^{k+1}, \textbf{a}_{l-1}^{k+1}&, \textbf{b}_{l-1}^{k+1}\big) - \mathcal{F}\big(\textbf{W}_{l}^{k+1}, \textbf{z}_{l-1}^{k+1}, \textbf{a}_{l-1}^{k+1}, \textbf{b}_{l}^{k+1}\big)\\ & \geq \frac{\zeta_l^{k+1}}{2} \norm{b_{l}^{k+1} - b_l^k}_2^2.\\
    \end{split}
\end{equation}   
\end{lemma}
Lemma \ref{lemma2} illustrates that the objective declines as all variables are updated. The following convergence properties hold concerning Assumption \ref{assumption1} and Lemma \ref{lemma2}. A similar idea has been applied in \cite{wang2022accelerated} to prove this Lemma, and thus it is omitted.  

\begin{lemma} \label{lemma3} (Objective Decrease \cite[Property 2]{wang2021convergent}), \\
    $\mathcal{F}(\textbf{W}^{k+1}, \textbf{z}^{k+1}, \textbf{a}^{k+1}, \textbf{b}^{k+1}) \leq \mathcal{F}(\textbf{W}^{k}, \textbf{z}^{k}, \textbf{a}^{k}, \textbf{b}^{k})$ for any $k \in \mathbb{N}$. Furthermore, $\mathcal{F}(\textbf{W}^{k}, \textbf{z}^{k}, \textbf{a}^{k}, \textbf{b}^{k}) \rightarrow \mathcal{F}^*$ as $k \rightarrow \infty$, ($\mathcal{F}^*$ is the convergent value of $\mathcal{F}$).
\end{lemma}
Lemma \ref{lemma3} depicts the decrease in the objective value during the iterations.
\begin{lemma} \label{lemma4}
    In AA-DLADMM algorithm, $\mathcal{F}(\textbf{W}^{k}, \textbf{z}^{k}, \textbf{a}^{k}, \textbf{b}^{k})$ is upper bounded for any $k \in \mathbb{N}$. Moreover, some scalars $M_W, M_z, M_a$, and $M_b$ exist such that $\norm{\textbf{W}^k} \leq M_W, \norm{\textbf{z}^k} \leq M_z, \norm{\textbf{a}^k} \leq M_a$, and $\norm{\textbf{b}^k} \leq M_b$, which implies that $(\textbf{W}^{k}, \textbf{z}^{k}, \textbf{a}^{k}, \textbf{b}^{k})$ are bounded.
\end{lemma}

This lemma shows that all variables and the objective are bounded and $\lim_{k \rightarrow \infty} \textbf{W}^{k+1} -\textbf{W}^k =0, \lim_{k \rightarrow \infty} \textbf{z}^{k+1} -\textbf{z}^k =0, \lim_{k \rightarrow \infty} \textbf{a}^{k+1} -\textbf{a}^k =0 $, and $\lim_{k \rightarrow \infty} \textbf{b}^{k+1} -\textbf{b}^k =0$. 
The reader can refer to \cite{wang2022accelerated} for the proof.

\begin{lemma} \label{lemma5}
    For any $k \in \mathbb{N}$, there exist a constant $C>0$, and $g_1^{k+1} \in \partial_{\textbf{W}^{k+1}} \mathcal{F} ((\textbf{W}^{k+1}, \textbf{z}^{k+1}, \textbf{a}^{k+1}, \textbf{b}^{k+1}))$ such that
    \begin{equation}\begin{split}
        \norm{g_1^{k+1}} \leq C &\bigg(\norm{\textbf{W}^{k+1} -\textbf{W}^k} + \norm{\textbf{z}^{k+1} -\textbf{z}^k} \\ &+  \norm{\textbf{b}^{k+1} -\textbf{b}^k}\bigg).
         \end{split}
    \end{equation}
\end{lemma}

Lemma \ref{lemma5} indicates that the subgradient of the objective is bounded regarding its variables. On this account, the subgradient converges to $0$; thereby, it converges to a stationary point \cite{wang2021convergent}. The proof is provided in Appendix \ref{proof lemma5}.
\begin{property} [Convergence of the residual $F$] \label{lemma6}
    Using (\ref{Bk-1 norm}), we obtain for an arbitrary fixed point $y\in X$ of $G$ that
    \begin{equation}\begin{split} \label{square}
        \norm{x_{k+1} -y}_2 &\leq \norm{x_k -y}_2 +\norm{B_k^{-1} F_k}_2\\
        & \leq \norm{x_{k} -y}_2 +C_G \norm{F_k}\\
        & \leq \norm{x_{k} -y}_2 +C_G d \bar{U}(n_{AA}+1)^{-(1+\varepsilon)}.
        \end{split}
    \end{equation}
    Then we obtain
    \begin{equation}\begin{split}
        \norm{x_k -y}_2 &\leq \norm{x_0 -y}_2 +C_G d \bar{U} \sum_{n_{AA}=0}^\infty (n_{AA}+1)^{-(1+\varepsilon)}\\& =E < \infty
        \end{split}
    \end{equation}
    Hence $\norm{x_k -y}_2$ stays bounded for $k \geq 0$. By taking the square from inequality (\ref{square}), we acquire 
    \begin{equation}\begin{split}
        \norm{x_{k+1} -y}_2^2 &\leq \norm{x_k -y}_2^2 + (C_G d \bar{U})^2 (n_{AA}+1)^{-(2+2\varepsilon)}\\
        &+ 2 C_G d E \bar{U}(n_{AA}+1)^{-(1+\varepsilon)}\\
    \end{split}
    \end{equation}
\end{property}
According to line 12 of Algorithm 1, we have $\lim_{k\rightarrow \infty} \norm{F_k}_2 =0$, and by defining $\varepsilon_k=(C_G d \bar{U})^2 (n_{AA}+1)^{-(2+2\varepsilon)}$, we obtain
\begin{equation}
    \norm{x_{k+1} -y}_2^2 \leq \norm{x_k -y}_2^2 + \varepsilon_k
\end{equation}
where $\varepsilon_k \geq 0$ and $\sum_{k=0}^\infty < \infty$.

We summarize the result of the aforementioned Lemmas and Property 1 in Theorem \ref{theorem 1}, which indicates the convergence of the proposed AA\_DLADMM algorithm using Anderson acceleration.

\begin{theorem}[Convergence under Anderson acceleration] \label{theorem 1}
   Suppose that Assumption \ref{assumption1} holds. Under the safeguarding step of Anderson acceleration, for any $\rho>0$ and $\epsilon>0$, Algorithm 1 converges to a stationary point $x^*$. That is, $0 \in \partial \mathcal{F}_{W^*} (\textbf{W}^*, \textbf{z}^*, \textbf{a}^*, \textbf{b}^*)$.
\end{theorem}

\begin{theorem}[r-linear Convergence Rate] \label{theorem2}
    In the AA-DLADMM algorithm, if $\mathcal{F}$ is a lower semicontinuous convex subanalytic function, then the convergence rate of ($\textbf{W}^k, \textbf{z}^k, \textbf{a}^k, \textbf{b}^k$) is r-liear.
    
\end{theorem}
The proof is given in Appendix \ref{proof theorem 2}.

\section{Experiments}\label{experiments}

In this section, we evaluate the proposed algorithm utilizing four benchmark datasets. The proposed algorithm's efficiency, convergence, and performance are compared with state-of-the-art approaches. All experiments are executed on a 64-bit machine with Intel(R) Iris(R) Xeon processor and NVIDIA GeForce RTX 3060 Ti.

\begin{table*}[h!]
\begin{center}
\caption{Statics of four benchmark datasets}
\label{table:statictis}
\begin{tabular}{p{3cm} P{3cm} P{3cm} P{3cm} P{2cm}}
\toprule
Dataset & Num Node & Num Edge &  Num Class & Num Feature\\                                   \\
\midrule
Cora          & 2708    & 5429     & 7     & 1433               \\ 
Pubmed           &19717    & 44338     & 3      & 500               \\ 
Citeseer         & 3327    & 4732     & 6      & 3703                  \\
Coauthor\_CS  & 18333 & 81894    & 15      & 6805                \\ 
\midrule
\end{tabular}
\end{center}
\end{table*}

\begin{table*}[h!]
\begin{center}
\caption{Hyperparameter setting on four datasets: they were chosen based on training performance}
\label{table1: hyperparam}
\begin{tabular}{p{2.5cm} P{2.5cm} P{2.5cm} P{2.5cm} P{2.5cm} P{2.5cm}}
\toprule
Method & Hyper\_parameters & Cora &  Pubmed & Citeseer & Coauthor\_CS\\                                   \\
\midrule
Proposed method          & $\rho$    & 1 $\times$ $10^{-4}$     & 0.045     & 1 $\times$ $10^{-3}$  & 7 $\times$ $10^{-4}$            \\ 
mDLAM          &$\rho$    & 1 $\times$ $10^{-3}$     & 0.01      & 5 $\times$ $10^{-3}$  &   1 $\times$ $10^{-4}$             \\ 
dlADMM         & $\rho$    & 1 $\times$ $10^{-6}$     & 1 $\times$ $10^{-6}$      & 1 $\times$ $10^{-6}$      & 1 $\times$ $10^{-6}$           \\
GD         & $\alpha$    & 0.01    & 0.01     & 0.01    & 0.01         \\
Adadelta         & $\alpha$    & 0.01    & 0.1     & 0.01    & 0.05         \\
Adagrad         & $\alpha$    & 5 $\times$ $10^{-3}$    & 5 $\times$ $10^{-3}$     & 0.01    & 5 $\times$ $10^{-3}$         \\
Adam         & $\alpha$    & 1 $\times$ $10^{-3}$    & 5 $\times$ $10^{-4}$     &  1 $\times$ $10^{-3}$    & 1 $\times$ $10^{-3}$         \\
\midrule
\end{tabular}
\end{center}
\end{table*}

\subsection{Experimental Setup}

In the following section, we explain the datasets and the structure of the networks utilized for our experiments.

\subsubsection{Dataset} 
In this experiment, we evaluate the performance of our model on four benchmark datasets. The datasets' statistics are shown in \hyperlink{thesentence}{Table \ref{table:statictis}} and are summarized below:
\begin{description}
    \item[Cora \cite{sen2008collective}] This dataset comprises 2708 machine learning publications categorized into one of seven classes. The citation networks contain 5429 links and 1433 specific words in the dictionary.
    \item[Pubmed \cite{sen2008collective}] The Pubmed dataset consists of 19717 scientific papers from PubMed databases concerning diabetes divided into one of three classes. The citation networks include 44338 links and 500 distinct words in the dictionary. 
    \item[Citeseer \cite{sen2008collective}] The Citeseer dataset contains 3312 scientific publications gathered from the \url{Tagged.com} social network website. It is classified into six classes and consists of 4732 links and 3703 unique words.
    \item[Coauthor CS \cite{shchur2018pitfalls}] The Coauthor CS dataset is derived from the Microsoft Academic Graph from the KDD Cup 2016 contest. The nodes represent the authors connected by an edge if they co-authored a publication. Here, the keywords of each author's papers are node features, and class labels denote each author's most active research areas. 
\end{description}

\subsubsection{Experimental Settings}

We set up a fully connected neural network as the basic model, which consists of three layers with 100 hidden units each. We considered the Rectified Linear Unit (ReLU) as a nonlinear activation function for each neuron. The cross-entropy with softmax was chosen as the loss function.  The number of the epoch was set to 200. 

In this study, the GD and its variant and ADMM are the state-of-the-art optimizers and thus carry out as comparison approaches. The full batch dataset trained the models. All parameters were selected on account of the maximum training accuracy. The baselines are outlined as follows:

\begin{description}
    \item[Gradient Descent (GD) \cite{bottou2010large}] The GD is the most recognized deep learning optimizer in which parameters are updated concerning their gradient. The convergence of this method has been substantially investigated in the literature \cite{ruder2016overview}.
    \item[Adaptive Gradient (Adagrad) \cite{duchi2011adaptive}] It is a modified variant of the gradient descent optimization algorithm that enables the learning rate to be automatically adapted to the parameters.
    \item[Adaptive learning rate (Adadelta) \cite{zeiler2012adadelta}] It is an extension of the Adagrad to address the sensitivity to hyperparameter choice.
    \item[Adaptive momentum estimation (Adam) \cite{kingma2014adam}] Adam is the most widespread deep learning optimization algorithm in generalizing proficiency. It approximates the first and second momentums to improve the biased gradient, thereby accelerating the convergence rate.
    \item[Alternating Direction Method of Multipliers (ADMM) \cite{taylor2016training}] ADMM is an algorithm to solve convex optimization problems by breaking the objective function into smaller subproblems. Then, they cooperated in obtaining global solutions. This method can be adaptable to large-scale datasets and allows parallel computations. 
\end{description}

The value of the hyperparameter used for all the comparison methods is depicted in \hyperlink{thesentence}{Table \ref{table1: hyperparam}}. For the proposed algorithm and mDLAM  method, the quadratic terms in Problem 3 are managed by $\rho$. In the dlADMM algorithm, $\rho$ regulates a linear constraint and $\alpha$ is a learning rate regarding the other gradient-based methods. To relax the inequality constraints at the early phase in the proposed algorithm and mDLAM, $\epsilon$ is set adaptively to $\epsilon^{k+1}=\max (\frac{\epsilon^k}{2}, 0.001)$ with $\epsilon^0 =100$.

\subsection{Experimental Results}
In this section, we analyze the experimental outcomes of the proposed algorithm against the other state-of-the-art methods. The following criteria are investigated for comparison. (1) Convergence: the correlation between the number of iterations and objective value. (2) Performance: the test accuracy of all comparison methods on four benchmark datasets. (3) Running time: averaged one epoch running time respecting two parameters: the value of $\rho$ and the number of hidden units. (4) Accuracy: the relationship between the maximum acquired test accuracy and the value of $m$ and $\rho$. (5) Memory: memory usage of the proposed algorithm and mDLAM on all datasets.

\subsubsection{Convergence}
\begin{figure*}[h!]
\captionsetup[subfigure]{justification=Centering}
\begin{subfigure}[t]{0.45\textwidth}
    \includegraphics[trim = 0mm 0mm 0mm 0mm, width=1\linewidth]{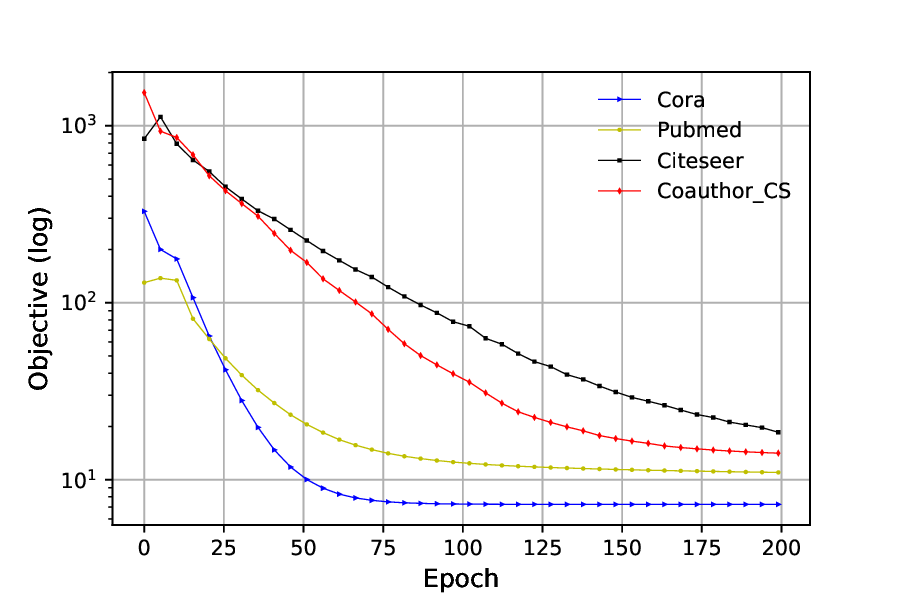}
    \caption{Objective Value}
\end{subfigure} 
\hfill
\begin{subfigure}[t]{0.45\textwidth}
    \includegraphics[trim = 0mm 0mm 0mm 0mm, width=1\linewidth]{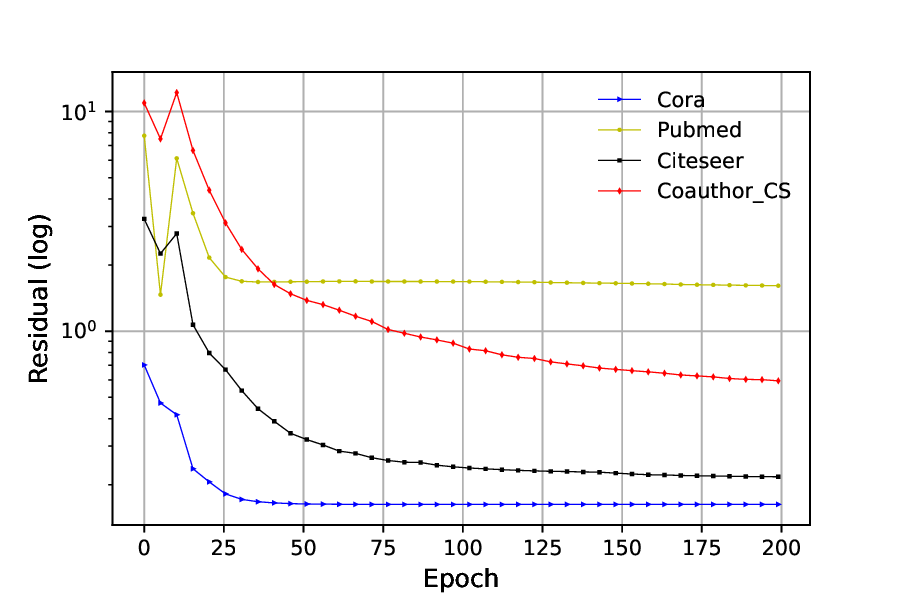}
    \caption{Residual}
\end{subfigure}
\caption{Convergence curves of the proposed algorithm on four datasets: they all converge linearly when the number of iterations exceeds 100 roughly in (a) and 75 in (b).}
\label{convergence}
\end{figure*}

\hyperlink{thesentence}{Fig. \ref{convergence}} depicts the convergence curves of the proposed algorithm on four datasets using the hyperparameters detailed in \hyperlink{thesentence}{Table \ref{table1: hyperparam}}. In \hyperlink{thesentence}{Fig. \ref{convergence}}(a) and (b), the X-axis denotes the number of epochs and the Y-axis represents the logarithm of the objective value and residual, respectively. On all datasets, both the objective value and residual decline monotonously, which validates the convergence property of the proposed algorithm. \hyperlink{thesentence}{Fig. \ref{convergence}}(a) illustrates the trend differences in the objective curves. In the beginning, the objective curves on the Cora and Pubmed datasets descend significantly and then attain the plateau after about 75 iterations, whilst the objective curves of the Coauthor CS reach the plateau when the iteration is roughly 150. The objective curve on the Citeseer maintains the downward trend over all iterations. Moreover, the residual and objective values on the Cora dataset are the lowest among other datasets at the end of the training. However, as shown in \hyperlink{thesentence}{Fig. \ref{convergence}}(b), the residual on the Pubmed dataset is around 1.6, which is five times more than other residuals of remaining datasets.

\subsubsection{Performance}
\begin{figure*}[h!]
\captionsetup[subfigure]{justification=Centering}
\begin{subfigure}[t]{0.45\textwidth}
    \includegraphics[trim = 0mm 0mm 0mm 0mm, width=1\linewidth]{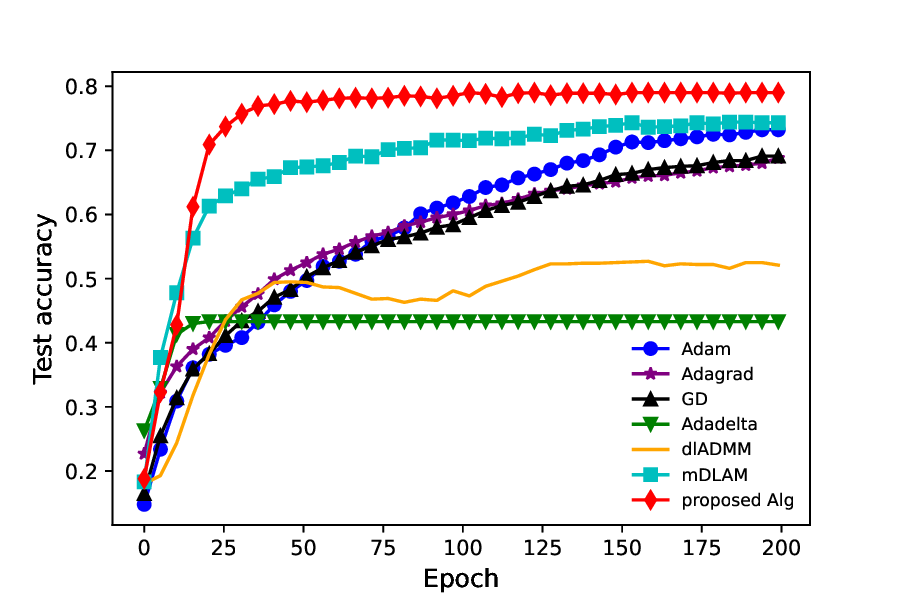}
    \caption{Cora}
\end{subfigure} 
\hfill
\begin{subfigure}[t]{0.45\textwidth}
    \includegraphics[trim = 0mm 0mm 0mm 0mm, width=1\linewidth]{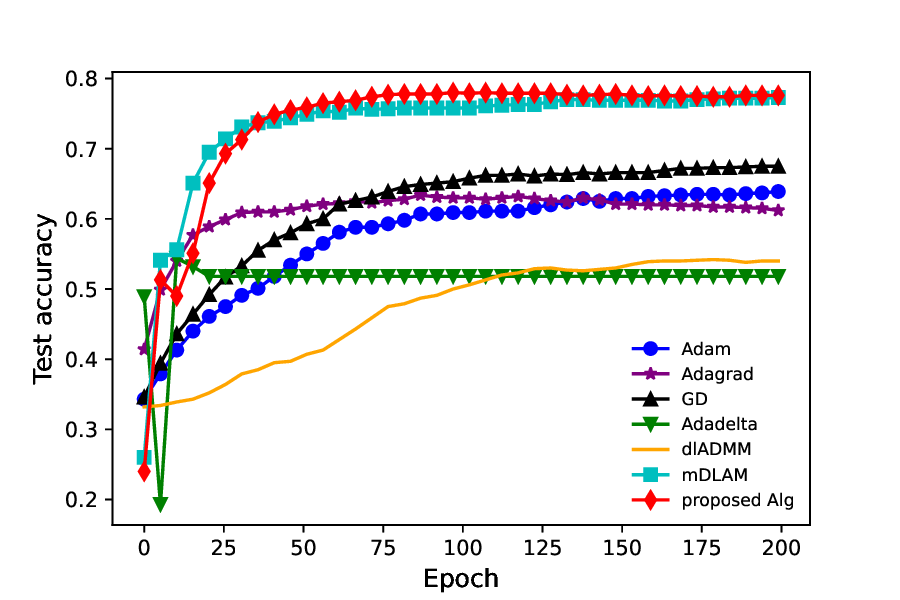}
    \caption{Pubmed}
\end{subfigure}

\begin{subfigure}[t]{0.45\textwidth}
    \includegraphics[trim = 0mm 0mm 0mm 0mm,width=1\linewidth]{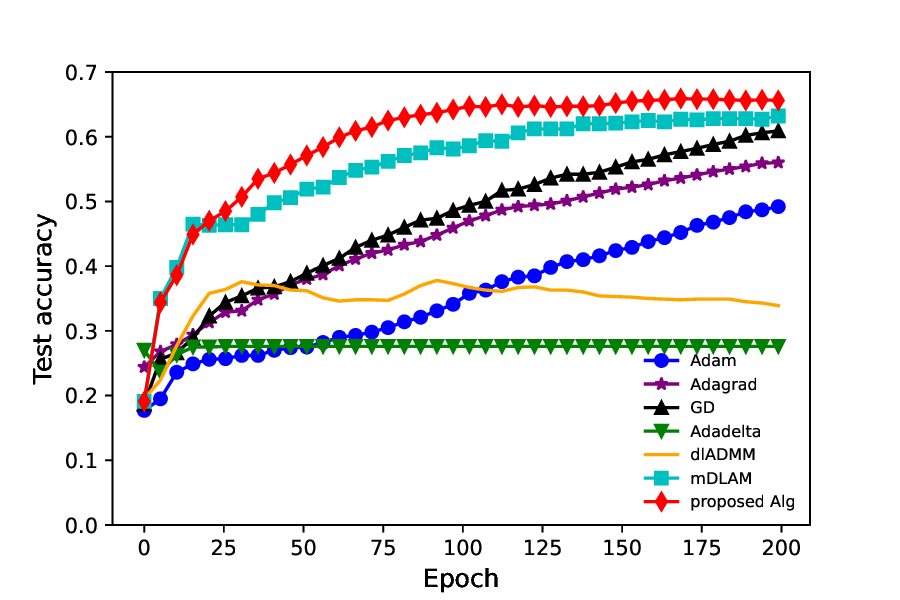}
    \caption{Citeseer}
\end{subfigure} 
\hfill
\begin{subfigure}[t]{0.45\textwidth}
    \includegraphics[trim = 0mm 0mm 0mm 0mm,width=1\linewidth]{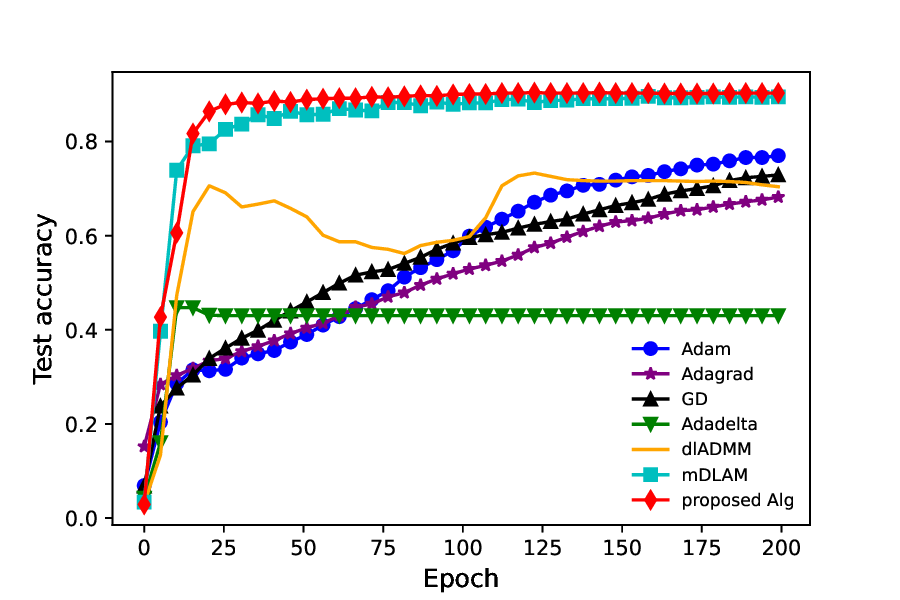}
    \caption{Coauthor\_CS}
\end{subfigure}

\caption{Performance of all methods for four datasets: the proposed method outperforms all other algorithms.}
\label{performance}
\end{figure*}

\hyperlink{thesentence}{Fig. \ref{performance}} shows the performance of the proposed method compared to the six state-of-the-art algorithms on four benchmark datasets. The $X$ and $Y$ axes denote the number of epochs and test accuracy, respectively. According to the experimental outcomes shown in \hyperlink{thesentence}{Fig. \ref{performance}}, the proposed algorithm not only acquired the highest peak accuracy but also achieved convergence rapidly as against other comparison methods. As an illustration, the proposed algorithm attained 70\% test accuracy in just 20 epochs on the Cora dataset, while mDLAM merely achieved 60\%, GD gained 38\% and Adadelta became flat around 43\%. In addition, \hyperlink{thesentence}{Fig. \ref{performance}}(a) depicts that the proposed algorithm greatly outperformed other approaches (approximately 80\%). For the Citeseer dataset, the proposed algorithm is the superior optimizer, followed by the mDLAM algorithm, which is 2.3\% lower in performance. In \hyperlink{thesentence}{Fig. \ref{performance}}(d), the curves of the proposed algorithm notably surged to around 85\% at the early stage and then increased steadily to 90\% or more. However, other comparison methods, including GD and Adam, acquired half of their test accuracy at the $25^{th}$ iteration (e.g., almost 40\%). The worst algorithm is Adadelta, whose curves mount the slowest: its performance on the Cora, Pubmed, Citeseer, and Coauthor CS is 35\%, 25\%, 37\%, and 53\% inferior to the proposed algorithm, respectively. The mDLAM approach is on par with the proposed algorithm in some cases: the test accuracy of mDLAM substantially reaches that of the proposed algorithm on the Pubmed and Coauthor CS datasets. Due to the observation of the experimental results, the curves associated with the Adagrad algorithm on the Pubmed dataset and dlADMM on the Citeseer dataset drop at the end of epochs.

\subsubsection{Running time}
Next, we investigate the relationship between our proposed algorithm's running time per iteration and two factors: the number of hidden units and the number of $\rho$. The running time is averaged by 200 iterations.

\begin{figure}[h!]
	\centering
		\includegraphics[trim = 0mm 0mm 0mm 0mm, width=1\linewidth]{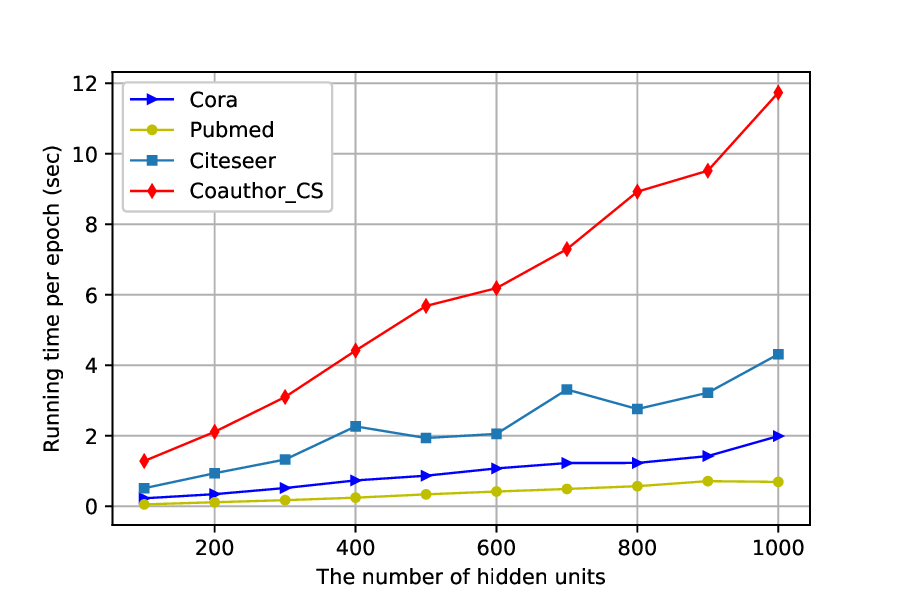}
	  \caption{The relationships between the running time per epoch (sec) and the number of hidden units}\label{neuron}
\end{figure}

Firstly, we analyze the effect of altering the number of hidden units on running time by increasing 100 neurons each time, ranging from 100 to 1000. \hyperlink{thesentence}{Fig. \ref{neuron}} shows the computational results for all four datasets. Overall, the running time increased linearly along with the growth of the number of neurons. However, there are some cases: for instance, the running time per epoch dropped from 2.265 to 1.935 when the number of hidden units increased from 400 to 500 on the Citeseer dataset. The running times are under 1 second per epoch for all datasets except Coauthor CS, which is marginally more than 1 second (1.282 s) when the number of neurons is 100. Nevertheless, the pace of increment differs on various datasets: the curve slope of the Coauthor CS soar sharply, which arrives at 12 seconds per epoch as the number of neurons is 1000, while the curves on the Pubmed and Cora rise subtly, which never exceed 1 and 2 seconds, respectively.  

\begin{figure}[h!]
	\centering
		\includegraphics[trim = 0mm 0mm 0mm 0mm, width=1\linewidth]{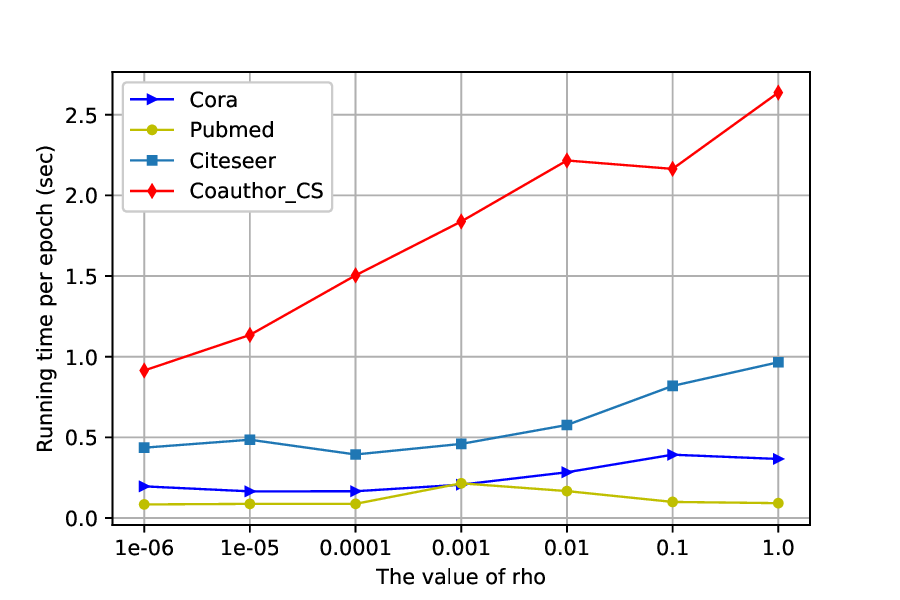}
	  \caption{The relationships between the running time per epoch (sec) and the value of $\rho$}\label{rho}
\end{figure}

Secondly, we explore the running time per iteration by changing the value of $\rho$ from $10^{-6}$ to 1 while the number of hidden units and $m$ are fixed to 100 and 8 for each layer. Generally, the running time increase as the result of enlarging the value of $\rho$, as depicted in \hyperlink{thesentence}{Fig. \ref{rho}}. The curve slope of Coauthor CS is superior, whereas the curve on the Pubmed, Cora, and Citeseer gradually climbs. Moreover, there exist some exceptions: the running time on Coauthor CS declines from 2.216 to 2.1164 seconds as the value of $\rho$ enlarges from $10^{-2}$ to $10^{-1}$, or the running time on the Pubmed drops from 0.215 to 0.092 seconds when $\rho$ rises from $10^{-3}$ to 1.

\subsubsection{Test accuracy}

In this experiment, we evaluate the impact of hyperparameters, namely, the value of $m$ and $\rho$ on test accuracy. \hyperlink{thesentence}{Table \ref{value rho}} describes the relation of test accuracy and the value of $\rho$, which alternated from $10^{-5}$ to $10^{-2}$. In general, the proper choice of $\rho$ plays a remarkable role in the test accuracy for all datasets. As an illustration, on the Pubmed dataset, when $\rho$ enlarged from $10^{-3}$ to $10^{-2}$, the performance enhanced to 0.776. However, the test accuracy on the Cora, Citeseer, and Coauthor CS datasets are comparably robust to the fluctuations of $\rho$ from $10^{-5}$ to $10^{-3}$. It should be mentioned that the performance of all datasets except Pubmed dropped dramatically when $\rho$ was set to $10^{-2}$, whereas the test accuracy of Pubmed improved the best to 0.776.

\hyperlink{thesentence}{Table \ref{value m}} demonstrates the correlation between test accuracy and the choice of $m$, which selects from {6, 8, 10, 12} on all datasets. Based on the obtained results, the performance is relatively resistant to the alteration of $m$. For instance, the test accuracy is near 0.90 regardless of the value of $m$ on the Coauthor CS dataset. For the Pubmed dataset, the choice of $m$ slightly affects test accuracy (around 0.12) when $m$ changed from 12 to 8. In all experimental results, $m=8$ achieved almost the best performance for all benchmark datasets. Comparing \hyperlink{thesentence}{Table \ref{value rho}} and \hyperlink{thesentence}{Table \ref{value m}}, the value of $\rho$ has a more significant impact on the test accuracy than that of $m$.    

\begin{table}[h!]
\caption{The relationship between test accuracy and the value of $\rho$}
    \begin{subtable}[h]{0.45\textwidth}
        \centering
        \begin{tabular}{l||*{5}{P{0.8cm}}}\hline
\backslashbox{$\rho$}{Epoch}
&\makebox[3em]{40}&\makebox[3em]{80}&\makebox[3em]{120}
&\makebox[3em]{160}&\makebox[3em]{200}\\\hline\hline
 $10^{-5}$ & 0.708 & 0.773 & 0.766 & 0.738 & 0.771 \\\hline
 $10^{-4}$ & 0.772 & 0.78 & 0.783 & 0.783 & 0.783\\\hline
 $10^{-3}$ & 0.662 & 0.717 & 0.743 & 0.748 & 0.749 \\\hline
 $10^{-2}$ & 0.298 & 0.271 & 0.27 & 0.263 & 0.258 \\\hline
\end{tabular}
\caption{Cora}
       \label{tab:week1}
    \end{subtable}
    \hfill
    \begin{subtable}[h]{0.45\textwidth}
        \centering
        \begin{tabular}{l||*{5}{P{0.8cm}}}\hline
\backslashbox{$\rho$}{Epoch}
&\makebox[3em]{40}&\makebox[3em]{80}&\makebox[3em]{120}
&\makebox[3em]{160}&\makebox[3em]{200}\\\hline\hline
 $10^{-5}$ & 0.407 & 0.405 & 0.413 & 0.413 & 0.413 \\\hline
 $10^{-4}$ & 0.4 & 0.399 & 0.4 & 0.401 & 0.402 \\\hline
 $10^{-3}$ & 0.413 & 0.413 & 0.413 & 0.413 & 0.413 \\\hline
 $10^{-2}$ & 0.752 & 0.782 & 0.779 & 0.776 & 0.776  \\\hline
\end{tabular}
\caption{Pubmed}
        \label{tab:week2}
     \end{subtable}
     \hfill
    \begin{subtable}[h]{0.45\textwidth}
        \centering
        \begin{tabular}{l||*{5}{P{0.8cm}}}\hline
\backslashbox{$\rho$}{Epoch}
&\makebox[3em]{40}&\makebox[3em]{80}&\makebox[3em]{120}
&\makebox[3em]{160}&\makebox[3em]{200}\\\hline\hline
 $10^{-5}$ & 0.694 & 0.621 & 0.642 & 0.656 & 0.663 \\\hline
 $10^{-4}$ & 0.637 & 0.656 & 0.66 & 0.657 & 0.658 \\\hline
 $10^{-3}$ & 0.544 & 0.63 & 0.648 & 0.657 & 0.657 \\\hline
 $10^{-2}$ & 0.262 & 0.237 & 0.233 & 0.233 & 0.233 \\\hline
\end{tabular}
\caption{Citeseer}
        \label{tab:week2}
     \end{subtable}

     \hfill
    \begin{subtable}[h]{0.45\textwidth}
        \centering
        \begin{tabular}{l||*{5}{P{0.8cm}}}\hline
\backslashbox{$\rho$}{Epoch}
&\makebox[2em]{40}&\makebox[2em]{80}&\makebox[2em]{120}
&\makebox[2em]{160}&\makebox[2em]{200}\\\hline\hline
 $10^{-5}$ & 0.885 & 0.896 & 0.894 & 0.894 & 0.894 \\\hline
 $10^{-4}$ & 0.882 & 0.894 & 0.901 & 0.904 & 0.904 \\\hline
 $10^{-3}$ & 0.826 & 0.865 & 0.866 & 0.874 & 0.876 \\\hline
 $10^{-2}$ & 0.523 & 0.576 & 0.591 & 0.608 & 0.608 \\\hline
\end{tabular}
\caption{Coathor\_CS}
        \label{tab:week2}
     \end{subtable}
     \label{value rho}
\end{table}

\begin{table}[h!]
\caption{The relationship between test accuracy and the choice of parameter m }
    \begin{subtable}[h]{0.45\textwidth}
        \centering
        \begin{tabular}{l||*{5}{P{0.8cm}}}\hline
\backslashbox{$m$}{Epoch}
&\makebox[3em]{40}&\makebox[3em]{80}&\makebox[3em]{120}
&\makebox[3em]{160}&\makebox[3em]{200}\\\hline\hline
 $m=6$ & 0.67 & 0.72 & 0.742 & 0.749 & 0.752 \\\hline
 $m=8$ & 0.772 & 0.78 & 0.805 & 0.783 & 0.783 \\\hline
 $m=10$ & 0.772 & 0.785 & 0.79 & 0.79 & 0.79\\\hline
 $m=12$ & 0.775 & 0.786 & 0.787 & 0.787 & 0.787 \\\hline
 
\end{tabular}
\caption{Cora}
       \label{tab:week1}
    \end{subtable}
    \hfill
    \begin{subtable}[h]{0.45\textwidth}
        \centering
        \begin{tabular}{l||*{5}{P{0.8cm}}}\hline
\backslashbox{$m$}{Epoch}
&\makebox[3em]{40}&\makebox[3em]{80}&\makebox[3em]{120}
&\makebox[3em]{160}&\makebox[3em]{200}\\\hline\hline
 $m=6$ & 0.753 & 0.775 & 0.778 & 0.779 & 0.778 \\\hline
 $m=8$ & 0.752 & 0.777 & 0.779 & 0.783 & 0.787 \\\hline
 $m=10$ & 0.754 & 0.779 & 0.775 & 0.776 & 0.778\\\hline
 $m=12$ & 0.201 & 0.591 & 0.638 & 0.653 & 0.658 \\\hline
\end{tabular}
\caption{Pubmed}
        \label{tab:week2}
     \end{subtable}
     \hfill
    \begin{subtable}[h]{0.45\textwidth}
        \centering
        \begin{tabular}{l||*{5}{P{0.8cm}}}\hline
\backslashbox{$m$}{Epoch}
&\makebox[3em]{40}&\makebox[3em]{80}&\makebox[3em]{120}
&\makebox[3em]{160}&\makebox[3em]{200}\\\hline\hline
 $m=6$ & 0.539 & 0.634 & 0.653 & 0.655 & 0.656 \\\hline
 $m=8$ & 0.544 & 0.63 & 0.648 & 0.657 & 0.657 \\\hline
 $m=10$ & 0.55 & 0.627 & 0.645 & 0.652 & 0.655 \\\hline
 $m=12$ & 0.636 & 0.649 & 0.655 & 0.657 & 0.658 \\\hline
\end{tabular}
\caption{Citeseer}
        \label{tab:week2}
     \end{subtable}

     \hfill
    \begin{subtable}[h]{0.45\textwidth}
        \centering
        \begin{tabular}{l||*{5}{P{0.8cm}}}\hline
\backslashbox{$m$}{Epoch}
&\makebox[3em]{40}&\makebox[3em]{80}&\makebox[3em]{120}
&\makebox[3em]{160}&\makebox[3em]{200}\\\hline\hline
 $m=6$ & 0.87 & 0.894 & 0.899 & 0.904 & 0.904 \\\hline
 $m=8$ & 0.882 & 0.894 & 0.901 & 0.904 & 0.904 \\\hline
 $m=10$ & 0.879 & 0.897 & 0.906 & 0.907 & 0.908 \\\hline
 $m=12$ & 0.879 & 0.901 & 0.907 & 0.908 & 0.907 \\\hline
\end{tabular}
\caption{Coathor\_CS}
        \label{tab:week2}
     \end{subtable}
     \label{value m}
\end{table}

\subsubsection{Memory analysis}

\begin{figure}[h!]
	\centering
		\includegraphics[trim = 0mm 0mm 0mm 0mm, width=1\linewidth]{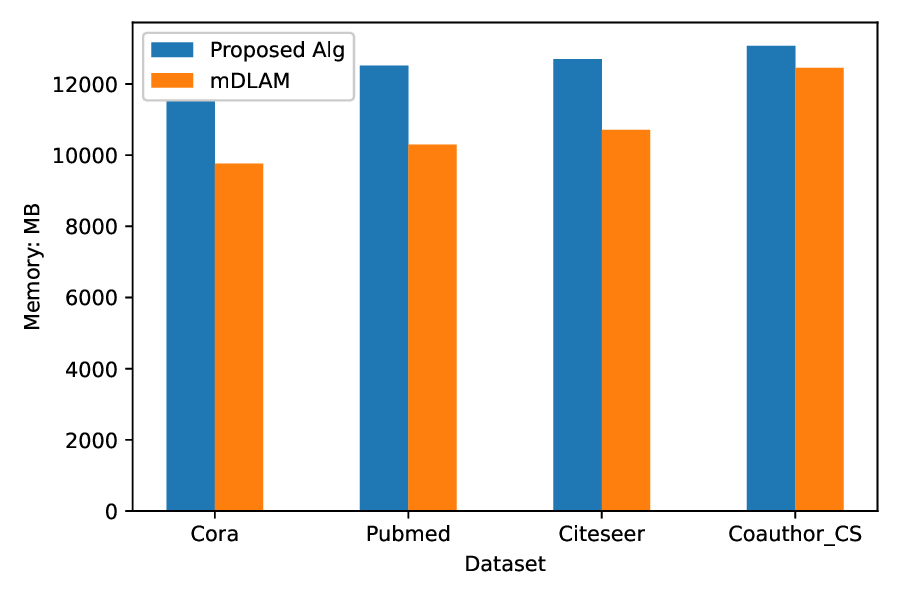}
	  \caption{Memory usage of the proposed Alg on all datasets versus mDLAM method}\label{memory}
\end{figure}

Finally, we compare the memory usage of our proposed algorithm with the mDLAM method on all datasets as illustrated in  \hyperlink{thesentence}{Fig. \ref{memory}}. We can realize the following deduction due to the observation of memory consumption. First, the Coauthor CS utilizes the most memory, around 13000MB, while the Cora consumes approximately 12200 MB in the proposed algorithm. Second, comparing the four datasets, the average difference in memory usage of our proposed algorithm and mDLAM is around 1800MB. Remarkably, our proposed algorithm occupies about 600MB more than mDLAM on the Coauthor CS dataset, albeit exerting more iteration for an update. Overall, the computational outcomes demonstrate that the proposed algorithm occupies more memory while reaching better accuracy and converging rapidly rather than other comparison methods.


\section{Conclusion} \label{conclusion}

The Alternating Direction Method of Multipliers (ADMM) is a superb alternative to Gradient Descent (GD) based methods for deep learning problems. In this study, we propose an efficient ADMM-based algorithm named Anderson Acceleration for Deep Learning Alternating Direction Method of Multipliers (AA-DLADMM). The Anderson acceleration approach is employed to elevate the convergence of the proposed AA-DLADMM method. Furthermore, the AA-DLADMM algorithm reaches a linear convergence, which is theoretically superior to other existing alternating optimizers. The experimental results on four benchmark datasets indicate not only the proficiency and convergence of our proposed AA-DLADMM algorithm but also outperform the other state-of-the-art optimizers.     

In future work, we may extend our AA-DLADMM from a fully connected neural network to train other prevalent neural networks, such as Convolution Neural Networks (CNN) and Recurrent Neural Networks (RNN).

\appendix

\section{Proof of Lemma \ref{lemma3}}
\label{proof lemma3}
By adding (\ref{f(w)}), (\ref{f(z)}), (\ref{f(a)}), and (\ref{f(b)}) for $l=1, \ldots L$, we acquire
\begin{equation}
\begin{split}
    &\mathcal{F} \left( \textbf{W}^k, \textbf{z}^k, \textbf{a}^k, \textbf{b}^k \right) - \mathcal{F} \left( \textbf{W}^{k+1}, \textbf{z}^{k+1}, \textbf{a}^{k+1}, \textbf{b}^{k+1} \right) \geq \\
    & \sum_{l=1}^L \Bigg( \frac{\beta_l^{k+1}}{2} \norm{W_{l}^{k+1} - W_l^k}_2^2 + \frac{\delta_l^{k+1}}{2} \norm{z_{l}^{k+1} - z_l^k}_2^2 \\
    &+ \frac{\zeta_l^{k+1}}{2} \norm{b_{l}^{k+1} - b_l^k}_2^2 \Bigg) + \sum_{l=1}^{L-1} \frac{\gamma_l^{k+1}}{2} \norm{a_{l}^{k+1} - a_l^k}_2^2.
    \end{split}
\end{equation}

Let $Cs=\min(\frac{\beta_l^{k+1}}{2}, \frac{\delta_l^{k+1}}{2}, \frac{\zeta_l^{k+1}}{2}, \frac{\gamma_l^{k+1}}{2})>0$, hence

\begin{equation} \label{fk-fk+1}
\begin{split}
    &\mathcal{F} \left( \textbf{W}^k, \textbf{z}^k, \textbf{a}^k, \textbf{b}^k \right) - \mathcal{F} \left( \textbf{W}^{k+1}, \textbf{z}^{k+1}, \textbf{a}^{k+1}, \textbf{b}^{k+1} \right) \geq \\
    & Cs \Bigg( \sum_{l=1}^L \Big(  \norm{W_{l}^{k+1} - W_l^k}_2^2 +  \norm{z_{l}^{k+1} - z_l^k}_2^2 \\
    &+  \norm{b_{l}^{k+1} - b_l^k}_2^2 \Big) + \sum_{l=1}^{L-1} \norm{a_{l}^{k+1} - a_l^k}_2^2 \Bigg)\\
    &= Cs \Big( \norm{W^{k+1} - W^k}_2^2 + \norm{z^{k+1} - z^k}_2^2\\
    &+ \norm{b^{k+1} - b^k}_2^2 + \norm{a^{k+1} - a^k}_2^2 \Big) \geq 0.
    \end{split}
\end{equation}

Owing to the boundedness of objectives and variables and the fact that the monotone sequence converges if it is bounded, then $\mathcal{F} \left( \textbf{W}^k, \textbf{z}^k, \textbf{a}^k, \textbf{b}^k \right)$ will converge.

\section{Proof of Lemma \ref{lemma5}}
\label{proof lemma5}
As we have
\begin{equation}
    \begin{split}
        \mathcal{F} \left(\textbf{W}_{l}^{k+1}, \textbf{z}_{l-1}^{k+1}, \textbf{a}_{l-1}^{k+1}, \textbf{b}_{l-1}^{k+1} \right) < \mathcal{F} \left(\textbf{W}_{l-1}^{k+1}, \textbf{z}_{l-1}^{k+1}, \textbf{a}_{l-1}^{k+1}, \textbf{b}_{l-1}^{k+1}\right)
    \end{split}
\end{equation}

We require a linear combination of $\norm{\textbf{W}^{k+1} - \textbf{W}^k}_2^2$,\\ $\norm{\textbf{z}^{k+1} - \textbf{z}^k}_2^2$, and $\norm{\textbf{b}^{k+1} - \textbf{b}^k}_2^2$ as an upper bound to prove Lemma \ref{lemma5}. First for $\textbf{W}^{k+1}$, according to the definition of $\mathcal{F}$ in problem 3, we have 
\begin{equation}
\begin{split}
    &\partial_{W^{k+1}} \mathcal{F}= \partial \Omega_{l} (W^{k+1})+ \nabla_{W_l^{k+1}} \phi (W_l^{k+1}, z_l^{k+1}, a_{l-1}^{k+1}, b_{l}^{k+1})\\
    &= \partial \Omega_{l} (W^{k+1})+ \nabla_{W_{l}^{k+1}} \phi (W_l^{k+1}, z_l^{k+1}, a_{l-1}^{k+1}, b_{l}^{k+1}) 
    \\& + \nabla_{\Bar{W}_{l}^{k+1}} \phi (\Bar{W}_l^{k+1}, z_l^{k+1}, a_{l-1}^{k}, b_{l}^{k}) + \theta_l^{k+1} (W_l^{k+1}-\Bar{W}_l^{k+1})\\
    & -\nabla_{\Bar{W}_{l}^{k+1}} \phi (\Bar{W}_l^{k+1}, z_l^{k+1}, a_{l-1}^{k}, b_{l}^{k}) - \theta_l^{k+1} (W_l^{k+1}-\Bar{W}_l^{k+1})\\
    & = \partial \Omega_{l} (W^{k+1}) + \nabla_{\Bar{W}_{l}^{k+1}} \phi (\Bar{W}_l^{k+1}, z_l^{k+1}, a_{l-1}^{k}, b_{l}^{k}) \\
    &+ \theta_l^{k+1}   (W_l^{k+1}-\Bar{W}_l^{k+1})+ \rho (W_l^{k+1} -\Bar{W}_l^{k+1}) a_{l-1}^{k+1} (a_{l-1}^{k+1})^\top \\
    & - \rho (z_l^{k+1} -z_l^k)(a_{l-1}^{k+1})^\top  - \rho (b_l^{k+1} -b_l^k)(a_{l-1}^{k+1})^\top \\
    &-\theta_l^{k+1} (W_l^{k+1}-\Bar{W}_l^{k+1}) 
\end{split}  
\end{equation}
By applying triangle inequality, we can obtain
\begin{equation}
    \begin{split}
        &\big\|\rho (W_l^{k+1} -\Bar{W}_l^{k+1}) a_{l-1}^{k+1} (a_{l-1}^{k+1})^\top - \rho (z_l^{k+1} -z_l^k)(a_{l-1}^{k+1})^\top\\ 
        &- \rho (b_l^{k+1} -b_l^k)(a_{l-1}^{k+1})^\top -\theta_l^{k+1} (W_l^{k+1}-\Bar{W}_l^{k+1}) \big\| \\
        & \leq \rho \norm{ (W_l^{k+1} -\Bar{W}_l^{k+1}) a_{l-1}^{k+1} (a_{l-1}^{k+1})^\top} + \rho  \big\|(z_l^{k+1} -z_l^k)\\
        & (a_{l-1}^{k+1})^\top\big\| + \rho \norm{(b_l^{k+1} -b_l^k)(a_{l-1}^{k+1})^\top} +\theta_l^{k+1} \norm{W_l^{k+1}-\Bar{W}_l^{k+1}}
    \end{split}
\end{equation}
Now according to Cauchy\_Schwartz inequality, we have
\begin{equation}\nonumber
    \begin{split}
        &\big\|\rho (W_l^{k+1} -\Bar{W}_l^{k+1}) a_{l-1}^{k+1} (a_{l-1}^{k+1})^\top - \rho (z_l^{k+1} -z_l^k)(a_{l-1}^{k+1})^\top\\ 
        & - \rho (b_l^{k+1} -b_l^k)(a_{l-1}^{k+1})^\top -\theta_l^{k+1} (W_l^{k+1}-\Bar{W}_l^{k+1})\big\| \\
        & \leq \rho \norm{ (W_l^{k+1} -\Bar{W}_l^{k+1})} \norm{a_{l-1}^{k+1}} \norm{(a_{l-1}^{k+1})^\top} + \rho \\& \norm{(z_l^{k+1} -z_l^k)} \norm{(a_{l-1}^{k+1})^\top} + \rho  \norm{(b_l^{k+1} -b_l^k)}  \norm{(a_{l-1}^{k+1})^\top} \\
        &+\theta_l^{k+1} \norm{W_l^{k+1}-\Bar{W}_l^{k+1}}
    \end{split}
\end{equation}
Based on Lemma \ref{lemma4}
\begin{equation} \nonumber
    \begin{split}
        &\leq (\rho M_a^2 +\theta_l^{k+1}) \norm{W_l^{k+1}-\Bar{W}_l^{k+1}} + \rho M_a \norm{(z_l^{k+1} -z_l^k)}\\
        &+ \rho M_a \norm{(b_l^{k+1} -b_l^k)}
    \end{split}
\end{equation}
By exerting Anderson Acceleration, we have
\begin{equation} \nonumber
    \begin{split}
        &\leq (\rho M_a^2 +\theta_l^{k+1}) \norm{W_l^{k+1}-(W_l^k - B_k^{-1}(W_l^k -W_l^{k+1}))} \\
        &+ \rho M_a \norm{(z_l^{k+1} -z_l^k)}+ \rho M_a \norm{(b_l^{k+1} -b_l^k)}\\
    \end{split}
\end{equation}
Now we apply triangle inequality to obtain
\begin{equation}
    \begin{split}
        & \leq (\rho M_a^2 +\theta_l^{k+1}) \norm{W_l^{k+1}-W_l^k }+ (\rho M_a^2 +\theta_l^{k+1})\\
        &\norm{B_k^{-1}(W_l^{k+1} -W_l^{k})} + \rho M_a \norm{(z_l^{k+1} -z_l^k)} \\
        &+ \rho M_a \norm{(b_l^{k+1} -b_l^k)}
    \end{split}
\end{equation}
Then due to (\ref{Bk-1 norm}) and Cauchy- Schwarz inequality, we acquire
\begin{equation}
    \begin{split}
        &\leq (\rho M_a^2 +\theta_l^{k+1}) \norm{W_l^{k+1}-W_l^k }+ (\rho M_a^2 +\theta_l^{k+1})\\
        &\left( \left( 3 \left(\frac{1+\overline{\theta}+\tau}{\tau} \right)^m -2 \right)^{n-1} \Big / \overline{\theta}^m \right)  \norm{W_l^{k+1}-W_l^k }\\
        &+ \rho M_a \norm{(z_l^{k+1} -z_l^k)} + \rho M_a \norm{(b_l^{k+1} -b_l^k)}
    \end{split}
\end{equation}
In addition, the optimality condition of (\ref{quadratic w}) results in
\begin{equation}
\begin{split}
        0 \in &\partial \Omega_{l} (W^{k+1}) + \nabla_{\Bar{W}_{l}^{k+1}} \phi (\Bar{W}_l^{k+1}, z_l^{k+1}, a_{l-1}^{k}, b_{l}^{k}) \\
        &+ \theta_l^{k+1}   (W_l^{k+1}-\Bar{W}_l^{k+1}).
\end{split}
\end{equation}
Let $CB= \left( 3 \left(\frac{1+\overline{\theta}+\tau}{\tau} \right)^m -2 \right)^{n-1} \Big / \overline{\theta}^m $. Thus, there exists $g_1^{k+1} \in \partial_{W_l^{k+1}} F$ such that
\begin{equation}
    \begin{split}
        \norm{g_1^{k+1}} &\leq (\rho M_a^2 +\theta_l^{k+1})(1+CB) \norm{W_l^{k+1}-W_l^k } \\
        &  + \rho M_a \norm{(z_l^{k+1} -z_l^k)} + \rho M_a \norm{(b_l^{k+1} -b_l^k)}.
    \end{split}
\end{equation}
Therefore, by considering $C=\max((\rho M_a^2 +\theta_l^{k+1})(1+CB), \rho M_a)$, we conclude that
\begin{equation} \label{g for Wlk+1}  \begin{split}
    \norm{g_1^{k+1}} \leq C \Big(&\norm{\textbf{W}^{k+1}-\textbf{W}^k }+\norm{(\textbf{z}^{k+1} -\textbf{z}^k)}+ \\
    & \norm{(\textbf{b}^{k+1} -\textbf{b}^k)} \Big).
    \end{split}
\end{equation}

On the other hand, when ($\Bar{W}_l^{k+1}= W_l^k$) we have
\begin{equation}
    \begin{split}
        &\big\|\rho (W_l^{k+1} -\Bar{W}_l^{k+1}) a_{l-1}^{k+1} (a_{l-1}^{k+1})^\top - \rho (z_l^{k+1} -z_l^k)(a_{l-1}^{k+1})^\top\\ 
        & - \rho (b_l^{k+1} -b_l^k)(a_{l-1}^{k+1})^\top -\theta_l^{k+1} (W_l^{k+1}-\Bar{W}_l^{k+1})\big\|  \\
        & \leq (\rho M_a^2 +\theta_l^{k+1}) \norm{W_l^{k+1}-\Bar{W}_l^{k+1}} + \rho M_a \norm{(z_l^{k+1} -z_l^k)}\\
        &+ \rho M_a \norm{(b_l^{k+1} -b_l^k)}\\
        &= (\rho M_a^2 +\theta_l^{k+1}) \norm{W_l^{k+1}- W_l^{k}} + \rho M_a \norm{(z_l^{k+1} -z_l^k)}\\
        &+ \rho M_a \norm{(b_l^{k+1} -b_l^k)} \hspace{0.3cm} (\text{Since} \Bar{W}_l^{k+1}= W_l^k).
    \end{split}
\end{equation}
Similarly, the optimality condition of (\ref{quadratic w}) yields
\begin{equation}
\begin{split}
        0 \in &\partial \Omega_{l} (W^{k+1}) + \nabla_{\Bar{W}_{l}^{k+1}} \phi (\Bar{W}_l^{k+1}, z_l^{k+1}, a_{l-1}^{k}, b_{l}^{k}) \\
        &+ \theta_l^{k+1}   (W_l^{k+1}-\Bar{W}_l^{k+1}).
\end{split}
\end{equation}
Then there exists $g_1^{k+1} \in \partial_{W_l^{k+1}} \mathcal{F}$ such that
\begin{equation} \label{g for wlk}
    \begin{split}
        \norm{g_1^{k+1}} &\leq (\rho M_a^2 +\theta_l^{k+1})\norm{W_l^{k+1}-W_l^k } \\
        &  + \rho M_a \norm{(z_l^{k+1} -z_l^k)} + \rho M_a \norm{(b_l^{k+1} -b_l^k)}
    \end{split}
\end{equation}
Thus, by combining (\ref{g for Wlk+1}) and (\ref{g for wlk}), we have
\begin{equation} \label{g1 for w}  \begin{split}
    \norm{g_1^{k+1}} \leq C \Big(&\norm{\textbf{W}^{k+1}-\textbf{W}^k }+\norm{(\textbf{z}^{k+1} -\textbf{z}^k)}+ \\
    & \norm{(\textbf{b}^{k+1} -\textbf{b}^k)} \Big).
    \end{split}
\end{equation}

\section{Proof of Theorem 2} \label{proof theorem 2}

In this proof, we use the Kurdyka-Lojasiewics (KL) property \cite{attouch2013convergence}. We say $f$ satisfies the KL property at $x^* \in dom(\partial f)$, if there exist $\eta \in (0,\infty]$, a neighborhood $U$ of $x^*$, and a continuous concave function $\varphi: [0,\eta) \rightarrow \mathbb{R}_+$ such that
\begin{inparaenum} [(i)]
    \item $\varphi (0)=0$, \hspace{0.17cm} \item $\varphi \in \mathbb{C}^1(0,\eta)$, \hspace{0.17cm} \item $\varphi'(s)>0, \forall s \in (0,\eta)$,
\end{inparaenum}
and (iv) for all $x \in U_\eta:=U \cap \{z \in \mathbb{R}^n, f(x^*)< f(z)<f(x^*)+\eta \}$, the KL inequality holds
\begin{equation} \label{definition kl}
    \varphi'(f(x)- f(x^*)) \hspace{0.1cm} \text{dist} (0, \partial f(x)) \geq 1.
\end{equation}

According to \cite[Definition F.1]{ouyang2020anderson}, $\varphi$ has the KL property at $x^*$ with exponent $\vartheta$, if $\varphi$ is chosen from Lojasiewics functions and satisfies $\varphi(x)=qx^{1-\vartheta}$ for $q>0$ and $\vartheta \in [0,1)$.

Firstly, we know that 
\begin{equation*} \begin{split}
    &\partial \mathcal{F} (\textbf{W}^{k+1}, \textbf{z}^{K+1}, \textbf{a}^{k+1}, \textbf{b}^{k+1})\\
    &=(\partial_{\textbf{W}^{k+1}} \mathcal{F}, \partial_{\textbf{z}^{k+1}} \mathcal{F}, \partial_{\textbf{a}^{k+1}} \mathcal{F}, \partial_{\textbf{b}^{k+1}} \mathcal{F} )
\end{split}
\end{equation*}
where $\partial_{\textbf{W}^{k+1}} \mathcal{F} =\{\partial_{W_l^{k+1}} \mathcal{F}\}_{l=1}^L$, $\partial_{\textbf{z}^{k+1}} \mathcal{F} =\{\partial_{z_l^{k+1}} \mathcal{F}\}_{l=1}^L$, $\partial_{\textbf{a}^{k+1}} \mathcal{F} =\{\partial_{a_l^{k+1}} \mathcal{F}\}_{l=1}^{L-1}$, $\partial_{\textbf{b}^{k+1}} \mathcal{F} =\{\partial_{b_l^{k+1}} \mathcal{F}\}_{l=1}^L$. To prove Theorem \ref{theorem2}, we require to provide an upper bound of $\partial_{\textbf{W}^{k+1}} \mathcal{F}$, $\partial_{\textbf{z}^{k+1}} \mathcal{F}$, $\partial_{\textbf{a}^{k+1}} \mathcal{F}$, and $\partial_{\textbf{b}^{k+1}} \mathcal{F}$ by a linear combination of $\norm{\textbf{W}^{k+1} - \textbf{W}^k}$, $\norm{\textbf{z}^{k+1} - \textbf{z}^k}$, $\norm{\textbf{a}^{k+1} - \textbf{a}^k}$, and $\norm{\textbf{b}^{k+1} - \textbf{b}^k}$.

For $b_l^{k+1}$,
\begin{equation*} \begin{split}
    \partial_{b_l^{k+1}} \mathcal{F} &= \nabla_{b_l^{k+1}} \phi (W_l^{k+1}, z_l^{k+1}, a_{l-1}^{k+1}, b_{l}^{k+1})\\
    &= \nabla_{b_l^{k+1}} \phi (W_l^{k+1}, z_l^{k+1}, a_{l-1}^{k+1}, b_{l}^{k+1}) \\
    &+ \nabla_{\Bar{b}_l^{k+1}} \phi (W_l^{k+1}, z_l^{k+1}, a_{l-1}^{k+1}, \Bar{b}_{l}^{k+1}) + \rho (b_l^{k+1} -\Bar{b}_l^{k+1})\\
    &- \rho (b_l^{k+1} -\Bar{b}_l^{k+1}) - \nabla_{\Bar{b}_l^{k+1}} \phi (W_l^{k+1}, z_l^{k+1}, a_{l-1}^{k+1}, \Bar{b}_{l}^{k+1})\\
    &= \rho (W_l^{k+1} a_{l-1}^{k+1} +b_l^{k+1} -z_l^{k+1}) + \rho (b_l^{k+1} -\Bar{b}_l^{k+1})\\
    &- \rho (b_l^{k+1} -\Bar{b}_l^{k+1}) -\rho (W_l^{k+1} a_{l-1}^{k+1} +\Bar{b}_l^{k+1} -z_l^{k+1}) \\
    &+\nabla_{\Bar{b}_l^{k+1}}
    \phi (W_l^{k+1}, z_l^{k+1}, a_{l-1}^{k+1}, \Bar{b}_{l}^{k+1}) \\
    &= \nabla_{\Bar{b}_l^{k+1}}
    \phi (W_l^{k+1}, z_l^{k+1}, a_{l-1}^{k+1}, \Bar{b}_{l}^{k+1}) + \rho (b_l^{k+1} -\Bar{b}_l^{k+1}) 
\end{split}
\end{equation*}

The optimality condition of (\ref{quadratibL}) yields
\begin{equation*}
    0 \in \nabla_{\Bar{b}_l^{k+1}}
    \phi (W_l^{k+1}, z_l^{k+1}, a_{l-1}^{k+1}, \Bar{b}_{l}^{k+1}) + \rho (b_l^{k+1} -\Bar{b}_l^{k+1}), 
\end{equation*}
Therefore, there exists $g_3^{k+1}=\partial_{b_l^{k+1}} \mathcal{F}$ such that
\begin{equation} \label{g for b}
    \norm{g_3^{k+1}}=0.
\end{equation}

For $z_l^{k+1} (l<L)$, since the solution to (\ref{quadratic z}) can be simplified as follows:
\begin{equation}
    z_l^{k+1} \leftarrow \Bar{z}_l^{k+1} - \nabla_{\Bar{z}_l^{k+1}} \phi /\rho.
\end{equation}
Therefore, we have
\begin{equation*}
    \begin{split}
        \partial_{z_l^{k+1}} \mathcal{F}&= \nabla_{z_l^{k+1}} \phi (W_l^{k+1}, z_l^{k+1}, a_{l-1}^{k+1}, b_{l}^{k+1})\\
        & =\nabla_{z_l^{k+1}} \phi (W_l^{k+1}, z_l^{k+1}, a_{l-1}^{k+1}, b_{l}^{k+1}) \\
        &-\nabla_{\Bar{z}_l^{k+1}} \phi (W_l^{k+1}, \Bar{z}_l^{k+1}, a_{l-1}^{k+1}, b_{l}^{k+1}) - \rho (z_l^{k+1} - \Bar{z}_l^{k+1})\\
        &=\rho (z_l^{k+1} -W_l^{k+1} a_{l-1}^{k+1} -b_l^{k+1}) - \rho (\Bar{z}_l^{k+1} -W_l^{k+1} a_{l-1}^{k+1}-b_l^{k+1})\\
        & -\rho (z_l^{k+1} - \Bar{z}_l^{k+1}) =0.
    \end{split}
\end{equation*}

For $z_L^{k+1}$,
\begin{equation*}
    \begin{split}
        \partial_{z_L^{k+1}} \mathcal{F}&= \nabla_{z_L^{k+1}} \phi (W_L^{k+1}, z_L^{k+1}, a_{L-1}^{k+1}, b_{L}^{k+1}) + \partial R(z_L^{k+1};y)\\
        & = \nabla_{z_L^{k+1}} \phi (W_L^{k+1}, z_L^{k+1}, a_{L-1}^{k+1}, b_{L}^{k+1}) + \partial R(z_L^{k+1};y)\\
        &+ \nabla_{\Bar{z}_L^{k+1}} \phi (W_L^{k+1}, \Bar{z}_L^{k+1}, a_{L-1}^{k+1}, b_{L}^{k+1}) + \rho (z_L^{k+1} -\Bar{z}_L^{k+1})\\
        & -\rho (z_L^{k+1} -\Bar{z}_L^{k+1}) - \nabla_{\Bar{z}_L^{k+1}} \phi (W_L^{k+1}, \Bar{z}_L^{k+1}, a_{L-1}^{k+1}, b_{L}^{k+1})  
    \end{split}
\end{equation*}
By the optimality condition of (\ref{quadratizL})
\begin{equation*}
    0 \in \partial R(z_L^{k+1};y)+ \nabla_{\Bar{z}_L^{k+1}} \phi (W_L^{k+1}, \Bar{z}_L^{k+1}, a_{L-1}^{k+1}, b_{L}^{k+1}) + \rho (z_L^{k+1} -\Bar{z}_L^{k+1}).
\end{equation*}
Then we obtain
\begin{equation*}
    \begin{split}
        \partial_{z_L^{k+1}} \mathcal{F}&= \nabla_{z_L^{k+1}} \phi (W_L^{k+1}, z_L^{k+1}, a_{L-1}^{k+1}, b_{L}^{k+1})\\
        &- \nabla_{\Bar{z}_L^{k+1}} \phi (W_L^{k+1}, \Bar{z}_L^{k+1}, a_{L-1}^{k+1}, b_{L}^{k+1}) -\rho (z_L^{k+1} -\Bar{z}_L^{k+1}) \\
        &\rho (z_L^{k+1} -W_L^{k+1} a_{L-1}^{k+1} -b_L^{k+1}) - \rho (\Bar{z}_L^{k+1} -W_L^{k+1} a_{L-1}^{k+1}-b_L^{k+1})\\
        & -\rho (z_L^{k+1} - \Bar{z}_L^{k+1}) =0.
    \end{split}
\end{equation*}
Thus, there exists $g_4^{k+1}=\partial_{z_l^{k+1}} \mathcal{F}$ such that
\begin{equation} \label{g for z}
    \norm{g_4^{k+1}}=0.
\end{equation}
For $a_l^{k+1}$
\begin{equation*}
    \begin{split}
        \partial_{a_l^{k+1}} \mathcal{F}&= \nabla_{a_l^{k+1}} \phi (W_{l+1}^{k+1}, z_{l+1}^{k+1}, a_{l}^{k+1}, b_{l+1}^{k+1})\\
    \end{split}
\end{equation*}
As the solution to (\ref{quadratic a}) can be expressed as follows:
\begin{equation}
    a_l^{k+1} \leftarrow \Bar{a}_l^{k+1} - \nabla_{\Bar{a}_l^{k+1}} \phi /\tau_l^{k+1},
\end{equation}
Hence we have
\begin{equation*}
    \begin{split}
        \partial_{a_l^{k+1}} \mathcal{F}&= \nabla_{a_l^{k+1}} \phi (W_{l+1}^{k+1}, z_{l+1}^{k+1}, a_{l}^{k+1}, b_{l+1}^{k+1})\\
        & - \nabla_{\Bar{a}_l^{k+1}} \phi(W_{l+1}^{k}, z_{l+1}^{k}, \Bar{a}_{l}^{k+1}, b_{l+1}^{k}) - \tau_l^{k+1}(a_l^{k+1}-\Bar{a}_l^{k+1})\\
        &= \rho (W_{l+1}^{k+1})^\top (W_{l+1}^{k+1} a_{l}^{k+1} +b_{l+1}^{k+1} -z_{l+1}^{k+1}) -\rho (W_{l+1}^{k})^\top\\
        &\hspace{0.4cm} (W_{l+1}^{k} \Bar{a}_{l}^{k+1} +b_{l+1}^{k} -z_{l+1}^{k}) - \tau_l^{k+1}(a_l^{k+1}-\Bar{a}_l^{k+1})\\
        & =\rho (W_{l+1}^{k+1})^\top W_{l+1}^{k+1}(a_l^{k+1} -\Bar{a}_l^{k+1}) + \rho (W_{l+1}^{k+1})^\top \\
        &\hspace{0.3cm} (W_{l+1}^{k+1} - W_{l+1}^k) \Bar{a}_l^{k+1} +\rho (W_{l+1}^{k+1} -W_{l+1}^k)^\top W_{l+1}^k \Bar{a}_l^{k+1}\\
        &- \rho (W_{l+1}^{k+1})^\top (z_{l+1}^{k+1}- z_{l+1}^k) -\rho (W_{l+1}^{k+1} - W_{l+1}^k)^\top z_{l+1}^k\\
        &+ \rho (W_{l+1}^{k+1})^\top (b_{l+1}^{k+1} -b_{l+1}^k) +\rho (W_{l+1}^{k+1} -W_{l+1}^k)^\top b_{l+1}^k\\
        &- \tau_{l}^{k+1} (a_l^{k+1}-\Bar{a}_l^{k+1})
    \end{split}
\end{equation*}
By applying Triangle inequality and Cauthy- Schwarz inequality, we acquire
\begin{equation*}
    \begin{split}
        \norm{\partial_{a_l^{k+1}} \mathcal{F}} &\leq \rho \norm{W_{l+1}^{k+1}} \norm{W_{l+1}^{k+1}} \norm{a_{l}^{k+1} -\Bar{a}_{l}^{k+1}} + \rho \norm{W_{l+1}^{k+1}}\\
        & \norm{W_{l+1}^{k+1} -W_{l+1}^k} \norm{\Bar{a}_{l}^{k+1}} +\rho \norm{W_{l+1}^{k+1} -W_{l+1}^k} \norm{W_{l+1}^k} \\
        &\norm{\Bar{a}_l^{k+1}} +\rho \norm{W_{l+1}^{k+1}} \norm{z_{l+1}^{k+1}- z_{l+1}^k} +\rho \norm{W_{l+1}^{k+1} - W_{l+1}^k}\\
        &\norm{z_{l+1}^k} + \rho \norm{W_{l+1}^{k+1}} \norm{b_{l+1}^{k+1} -b_{l+1}^k} +\rho \norm{W_{l+1}^{k+1} -W_{l+1}^k}\\
        & \norm{b_{l+1}^k} + \tau_{l}^{k+1} \norm{a_l^{k+1}-\Bar{a}_l^{k+1}}\\
        (\text{Lemma \ref{lemma4}})\\
        & \leq \rho M_W^2 \norm{a_{l}^{k+1} -\Bar{a}_{l}^{k+1}} +\rho M_W \norm{W_{l+1}^{k+1} -W_{l+1}^k} M_a \\
        & + \rho \norm{W_{l+1}^{k+1} -W_{l+1}^k} M_W M_a +\rho M_W \norm{z_{l+1}^{k+1}- z_{l+1}^k}\\
        &+ \rho \norm{W_{l+1}^{k+1} - W_{l+1}^k} M_z +\rho M_W \norm{b_{l+1}^{k+1} -b_{l+1}^k}\\
        &+ \rho \norm{W_{l+1}^{k+1} -W_{l+1}^k} M_b + \tau_{l}^{k+1} \norm{a_l^{k+1}-\Bar{a}_l^{k+1}}\\
        & \leq (\rho M_W^2 +\tau_{l}^{k+1})\norm{a_{l}^{k+1} -\Bar{a}_{l}^{k+1}} + (2\rho M_W M_a +\rho M_z \\
        &+\rho M_b) \norm{W_{l+1}^{k+1} - W_{l+1}^k} + \rho M_W \norm{z_{l+1}^{k+1}- z_{l+1}^k}\\
        & + \rho M_W \norm{b_{l+1}^{k+1} -b_{l+1}^k}
    \end{split}
\end{equation*}
Based on the Anderson acceleration
\begin{equation*}
    \begin{split}
        \norm{\partial_{a_l^{k+1}} \mathcal{F}} &\leq (\rho M_W^2 +\tau_{l}^{k+1})\norm{a_{l}^{k+1} -\left(a_l^k -B_k^{-1} \left(a_l^k -a_{l}^{k+1}\right)\right)}\\
        &+(2\rho M_W M_a +\rho M_z +\rho M_b) \norm{W_{l+1}^{k+1} - W_{l+1}^k} \\
        &+ \rho M_W \norm{z_{l+1}^{k+1}- z_{l+1}^k} + \rho M_W \norm{b_{l+1}^{k+1} -b_{l+1}^k}\\
        &\leq (\rho M_W^2 +\tau_{l}^{k+1}) \norm{a_{l}^{k+1} -a_l^k} +(\rho M_W^2 +\tau_{l}^{k+1}) \\
        & \norm{B_k^{-1} \left(a_{l}^{k+1} -a_l^k\right)}+ (2\rho M_W M_a +\rho M_z +\rho M_b)\\
        &\norm{W_{l+1}^{k+1} - W_{l+1}^k} + \rho M_W \norm{z_{l+1}^{k+1}- z_{l+1}^k} + \rho M_W\\
        &\norm{b_{l+1}^{k+1} -b_{l+1}^k} (\text{Triangle inequality})\\
        &\leq (\rho M_W^2 +\tau_{l}^{k+1}) \norm{a_{l}^{k+1} -a_l^k} +(\rho M_W^2 +\tau_{l}^{k+1}) CB\\
        & \norm{a_{l}^{k+1} -a_l^k} +(2\rho M_W M_a +\rho M_z +\rho M_b) \\
        &\norm{W_{l+1}^{k+1} - W_{l+1}^k} + \rho M_W \norm{z_{l+1}^{k+1}- z_{l+1}^k} + \rho M_W\\
        &\norm{b_{l+1}^{k+1} -b_{l+1}^k} ((\text{\ref{Bk-1 norm})  and Cauchy- Schwarz inequality})
    \end{split}
\end{equation*}
Thus, there exists $g_5^{k+1} \in \partial_{a_l^{k+1}} \mathcal{F}$ such that
\begin{equation} \label{g5 abar}
    \begin{split}
        \norm{g_5^{k+1}} &\leq (\rho M_W^2 +\tau_{l}^{k+1})(1+CB) \norm{a_{l}^{k+1} -a_l^k}\\
        &+(2\rho M_W M_a +\rho M_z +\rho M_b) \norm{W_{l+1}^{k+1} - W_{l+1}^k}\\
        &+ \rho M_W \norm{z_{l+1}^{k+1}- z_{l+1}^k} + \rho M_W \norm{b_{l+1}^{k+1} -b_{l+1}^k}.
    \end{split}
\end{equation}
Otherwise if ($\Bar{a}_l^{k+1}=a_l^k$), then we have
\begin{equation} \label{g5 a}
    \begin{split}
        \norm{g_5^{k+1}} &\leq (\rho M_W^2 +\tau_{l}^{k+1})\norm{a_{l}^{k+1} -a_l^k}\\
        &+(2\rho M_W M_a +\rho M_z +\rho M_b) \norm{W_{l+1}^{k+1} - W_{l+1}^k}\\
        &+ \rho M_W \norm{z_{l+1}^{k+1}- z_{l+1}^k} + \rho M_W \norm{b_{l+1}^{k+1} -b_{l+1}^k}.
    \end{split}
\end{equation}
Therefore, by defining $C_1= \max((\rho M_W^2 +\tau_{l}^{k+1})(1+CB), (2\rho M_W M_a +\rho M_z +\rho M_b), \rho M_W )$ and combining (\ref{g5 abar}) and (\ref{g5 a}), we infer that
\begin{equation} \label{g for a}
    \begin{split}
        \norm{g_5^{k+1}} &\leq C_1 \bigg(\norm{\textbf{a}^{k+1} -\textbf{a}^k} +\norm{\textbf{W}^{k+1} - \textbf{W}^k}\\
        & +\norm{\textbf{z}^{k+1}- \textbf{z}^k} + \norm{\textbf{b}^{k+1} -\textbf{b}^k}\bigg).
    \end{split}
\end{equation}

According to Lemma \ref{lemma5}, (\ref{g for b}), (\ref{g for z}), and (\ref{g for a}), we prove that there exists $g^{k+1} \in \partial \mathcal{F} (\textbf{W}^{k+1}, \textbf{z}^{K+1}, \textbf{a}^{k+1}, \textbf{b}^{k+1})
    =(\partial_{\textbf{W}^{k+1}} \mathcal{F}, \partial_{\textbf{z}^{k+1}} \mathcal{F}, \partial_{\textbf{a}^{k+1}} \mathcal{F}, \partial_{\textbf{b}^{k+1}} \mathcal{F} )$ and $C_2=\max(C, C_1, \rho)$ such that
\begin{equation} \label{g for dist}
    \begin{split}
        \norm{g^{k+1}} &\leq C_2 \bigg(\norm{\textbf{a}^{k+1} -\textbf{a}^k} +\norm{\textbf{W}^{k+1} - \textbf{W}^k}\\
        & +\norm{\textbf{z}^{k+1}- \textbf{z}^k} + \norm{\textbf{b}^{k+1} -\textbf{b}^k} \bigg).
    \end{split}
\end{equation}
Finally, we prove the convergence rate of the proposed algorithm by KL property. Let $\varkappa$ and $\eta$ be the constants used in the KL property. By Lemma \ref{lemma3}, $\mathcal{F}(\textbf{W}^{k}, \textbf{z}^{k}, \textbf{a}^{k}, \textbf{b}^{k}) \rightarrow \mathcal{F}^*$, then for any $\eta>0$ and $k \geq k'$, we have $\mathcal{F}^* < \mathcal{F}(\textbf{W}^{k}, \textbf{z}^{k}, \textbf{a}^{k}, \textbf{b}^{k}) <\mathcal{F}^* +\eta$. Define $\varkappa_k =\varphi (\mathcal{F}(\textbf{W}^{k}, \textbf{z}^{k}, \textbf{a}^{k}, \textbf{b}^{k}) -\mathcal{F}^* )$. We employed the concavity property of $\varkappa$ for $k \geq k'$ as follows
\begin{equation}
    \begin{split}
        \varkappa_k -\varkappa_{k+1} &\geq \varphi' (\varkappa_k) \big(\mathcal{F}(\textbf{W}^{k}, \textbf{z}^{k}, \textbf{a}^{k}, \textbf{b}^{k}) - \mathcal{F}(\textbf{W}^{k+1}, \textbf{z}^{k+1}, \textbf{a}^{k+1}\\
        &, \textbf{b}^{k+1})\big)\\
        & \geq \frac{\mathcal{F}(\textbf{W}^{k}, \textbf{z}^{k}, \textbf{a}^{k}, \textbf{b}^{k}) - \mathcal{F}(\textbf{W}^{k+1}, \textbf{z}^{k+1}, \textbf{a}^{k+1}, \textbf{b}^{k+1})}{\text{dist} \left(0, \partial \mathcal{F}(\textbf{W}^{k}, \textbf{z}^{k}, \textbf{a}^{k}, \textbf{b}^{k})\right)}
    \end{split}
\end{equation}

Then, we require to obtain $\text{dist} \left(0, \partial \mathcal{F}(\textbf{W}^{k}, \textbf{z}^{k}, \textbf{a}^{k}, \textbf{b}^{k})\right)$. Due to (\ref{g for dist}), we have
\begin{equation} \label{dist}
    \begin{split}
        \text{dist} \big(0, &\partial \mathcal{F}(\textbf{W}^{k}, \textbf{z}^{k}, \textbf{a}^{k}, \textbf{b}^{k})\big) = \norm{g^{k+1}} \leq   C_2 \bigg(\norm{\textbf{a}^{k+1} -\textbf{a}^k} \\
        &+\norm{\textbf{W}^{k+1} - \textbf{W}^k}
         +\norm{\textbf{z}^{k+1}- \textbf{z}^k} + \norm{\textbf{b}^{k+1} -\textbf{b}^k} \bigg).
    \end{split}
\end{equation}
Now according to Lemma \ref{lemma3} and the obtained result in (\ref{fk-fk+1}), we have 
\begin{equation} \label{bound fk}
    \begin{split}
        &F \left( \textbf{W}^k, \textbf{z}^k, \textbf{a}^k, \textbf{b}^k \right) - F \left( \textbf{W}^{k+1}, \textbf{z}^{k+1}, \textbf{a}^{k+1}, \textbf{b}^{k+1} \right) \geq \\
        &Cs \Big( \norm{W^{k+1} - W^k}^2 + \norm{z^{k+1} - z^k}^2
       + \norm{b^{k+1} - b^k}^2 \\
       &+ \norm{a^{k+1} - a^k}^2 \Big) . 
    \end{split}
\end{equation}
Then by (\ref{dist}) and (\ref{bound fk})
\begin{equation}
    \begin{split}
       & \varkappa_k -\varkappa_{k+1} \geq \frac{Cs}{C_2} \\
       & \left( \frac{ \norm{W^{k+1} - W^k}^2 + \norm{z^{k+1} - z^k}^2
       + \norm{b^{k+1} - b^k}^2 + \norm{a^{k+1} - a^k}^2 }{ \norm{\textbf{a}^{k+1} -\textbf{a}^k} 
    +\norm{\textbf{W}^{k+1} - \textbf{W}^k}
         +\norm{\textbf{z}^{k+1}- \textbf{z}^k} + \norm{\textbf{b}^{k+1} -\textbf{b}^k} }\right)\\
         &(\text{By Mean inequality}) \\
         &\frac{Cs \left(\norm{W^{k+1} - W^k} + \norm{z^{k+1} - z^k}
       + \norm{b^{k+1} - b^k} + \norm{a^{k+1} - a^k}\right)^2}{4C_2 \left(\norm{\textbf{a}^{k+1} -\textbf{a}^k} 
    +\norm{\textbf{W}^{k+1} - \textbf{W}^k}
         +\norm{\textbf{z}^{k+1}- \textbf{z}^k} + \norm{\textbf{b}^{k+1} -\textbf{b}^k}\right)} 
    \end{split}
\end{equation}
Let $\Bar{C}=\frac{Cs}{4 C_2}$, then we have
\begin{equation} \label{varkappa}
\begin{split}
    \varkappa_k -\varkappa_{k+1} \geq \Bar{C} \bigg( \norm{W^{k+1} - W^k} + \norm{z^{k+1} - z^k} + \norm{b^{k+1} - b^k} \\+ \norm{a^{k+1} - a^k} \bigg)
    \end{split}
\end{equation}
Now by summing (\ref{varkappa}) from $k'$ to $\infty$
\begin{equation}
    \begin{split}
        \varkappa_{k'} \geq \Bar{C} \sum_{k=k'}^{\infty} \bigg( \norm{W^{k+1} - W^k} + \norm{z^{k+1} - z^k} \\+ \norm{b^{k+1} - b^k} + \norm{a^{k+1} - a^k} \bigg).
    \end{split}
\end{equation}
It should be mentioned that $\varkappa_k$ is positive and monotone decreasing. Derived from Lemma \ref{lemma4}, every term in this infinite series converges to zero since the total of this infinite series is finite. This indicates that $\lim_{k \rightarrow \infty} \textbf{W}^{k+1} -\textbf{W}^k =0, \lim_{k \rightarrow \infty} \textbf{z}^{k+1} -\textbf{z}^k =0, \lim_{k \rightarrow \infty} \textbf{a}^{k+1} -\textbf{a}^k =0 $, and $\lim_{k \rightarrow \infty} \textbf{b}^{k+1} -\textbf{b}^k =0$. Now, we require to satisfy that $\mathcal{F}(\textbf{W}^{k}, \textbf{z}^{k}, \textbf{a}^{k}, \textbf{b}^{k})$ has the KL property with the exponent $\vartheta \in (0,\frac{1}{2}]$. By assuming $\varphi(x)=q x^{1-\vartheta}$ and the KL property's definition (\ref{definition kl}) and (\ref{dist})
\begin{equation} 
    \begin{split}
         &C_2 \left(q (1-\vartheta)\right)\bigg(\norm{\textbf{a}^{k+1} -\textbf{a}^k} 
        +\norm{\textbf{W}^{k+1} - \textbf{W}^k}
         +\norm{\textbf{z}^{k+1}- \textbf{z}^k} \\
         &+ \norm{\textbf{b}^{k+1} -\textbf{b}^k} \bigg) \geq 
         \left(q (1-\vartheta)\right)\text{dist} \big(0, \partial \mathcal{F}(\textbf{W}^{k}, \textbf{z}^{k}, \textbf{a}^{k}, \textbf{b}^{k})\big) \\
         & \geq \left(\mathcal{F}(\textbf{W}^{k}, \textbf{z}^{k}, \textbf{a}^{k}, \textbf{b}^{k}) - \mathcal{F}(\textbf{W}^{k+1}, \textbf{z}^{k+1}, \textbf{a}^{k+1}, \textbf{b}^{k+1})\right)^\vartheta.
    \end{split}
\end{equation}
Thus we obtain
\begin{equation}
    \begin{split}
        &\varkappa_k= q\left(\mathcal{F}(\textbf{W}^{k}, \textbf{z}^{k}, \textbf{a}^{k}, \textbf{b}^{k}) - \mathcal{F}(\textbf{W}^{k+1}, \textbf{z}^{k+1}, \textbf{a}^{k+1}, \textbf{b}^{k+1})\right)^{1-\vartheta}\\
        &\leq q\Bigg( C_2 \left(q (1-\vartheta)\right) \bigg(\norm{\textbf{a}^{k+1} -\textbf{a}^k} 
        +\norm{\textbf{W}^{k+1} - \textbf{W}^k}
         \\&+\norm{\textbf{z}^{k+1}- \textbf{z}^k} 
         + \norm{\textbf{b}^{k+1} -\textbf{b}^k} \bigg) \Bigg)^{\frac{1-\vartheta}{\vartheta}} 
    \end{split}
\end{equation}
It is clear that $\Bigg( C_2 \left(q (1-\vartheta)\right) \bigg(\norm{\textbf{a}^{k+1} -\textbf{a}^k} 
        +\norm{\textbf{W}^{k+1} - \textbf{W}^k}
         +\norm{\textbf{z}^{k+1}- \textbf{z}^k} 
         + \norm{\textbf{b}^{k+1} -\textbf{b}^k} \bigg) \Bigg)^{\frac{1-\vartheta}{\vartheta}} $ is monotone increasing when $\vartheta \in (0,\frac{1}{2}]$ if $\bigg(C_2 q \big(\norm{\textbf{a}^{k+1} -\textbf{a}^k} 
        +\norm{\textbf{W}^{k+1} - \textbf{W}^k}
         +\norm{\textbf{z}^{k+1}- \textbf{z}^k} 
         + \norm{\textbf{b}^{k+1} -\textbf{b}^k} \big) \bigg)<1$. Therefore, for any $k\geq k'$ we obtain
\begin{equation}
\begin{split}
    \varkappa_k \leq& C_3 \bigg(\norm{\textbf{a}^{k+1} -\textbf{a}^k} 
        +\norm{\textbf{W}^{k+1} - \textbf{W}^k}
         +\norm{\textbf{z}^{k+1}- \textbf{z}^k} \\
         &+ \norm{\textbf{b}^{k+1} -\textbf{b}^k} \bigg)
\end{split}
\end{equation}
where $C_3$ is constant. Then, we can acquire the following outcome similar to the previous proof.
\begin{equation} \label{sum k}
    \begin{split}
        \varkappa_k -\varkappa_{k'} \geq C_4 &\bigg(\norm{\textbf{a}^{k+1} -\textbf{a}^k}  +\norm{\textbf{W}^{k+1} - \textbf{W}^k}
         +\norm{\textbf{z}^{k+1}- \textbf{z}^k} \\
         &+ \norm{\textbf{b}^{k+1} -\textbf{b}^k} \bigg)
    \end{split}
\end{equation}
Then by summing (\ref{sum k}) from $k'$ to $\infty$, we have
\begin{equation}
    \begin{split}
        \varkappa_{k'} \geq C_4 \sum_{k=k'}^\infty & \bigg(\norm{\textbf{a}^{k+1} -\textbf{a}^k}  +\norm{\textbf{W}^{k+1} - \textbf{W}^k}
         +\norm{\textbf{z}^{k+1}- \textbf{z}^k} \\
         &+ \norm{\textbf{b}^{k+1} -\textbf{b}^k} \bigg)
    \end{split}
\end{equation}
Thus
\begin{equation}
    \begin{split}
        & C_3 \bigg(\norm{\textbf{a}^{k'+1} -\textbf{a}^{k'}} 
        +\norm{\textbf{W}^{k'+1} - \textbf{W}^{k'}}
         +\norm{\textbf{z}^{k'+1}- \textbf{z}^{k'}} \\
         &+ \norm{\textbf{b}^{k'+1} -\textbf{b}^{k'}} \bigg) \geq C_4 \sum_{k=k'}^\infty \bigg(\norm{\textbf{a}^{k+1} -\textbf{a}^k}  +\norm{\textbf{W}^{k+1} - \textbf{W}^k}\\
         &+\norm{\textbf{z}^{k+1}- \textbf{z}^k} 
         + \norm{\textbf{b}^{k+1} -\textbf{b}^k} \bigg)
    \end{split}
\end{equation}
Let $\mathcal{M}_k= \sum_{i=k}^\infty \bigg(\norm{\textbf{a}^{i+1} -\textbf{a}^i}  +\norm{\textbf{W}^{i+1} - \textbf{W}^i}
         +\norm{\textbf{z}^{i+1}- \textbf{z}^i} 
         + \norm{\textbf{b}^{i+1} -\textbf{b}^i} \bigg)$, then we have
\begin{equation}
\begin{split}
    C_3 (\mathcal{M}_{k'} -\mathcal{M}_{k'+1}) \geq C_4 \mathcal{M}_{k'} \\
    \Rightarrow \mathcal{M}_{k'+1} \leq \frac{C_3 -C_4}{C_3} \mathcal{M}_{k'}
\end{split}
\end{equation}
Therefore, it is deduced that $\mathcal{M}_{k}$ converge q-linearly and attain the r-linear convergence of $\bigg(\norm{\textbf{a}^{k+1} -\textbf{a}^k}  +\norm{\textbf{W}^{k+1} - \textbf{W}^k}
         +\norm{\textbf{z}^{k+1}- \textbf{z}^k} 
         + \norm{\textbf{b}^{k+1} -\textbf{b}^k} \bigg)$.

\clearpage




\bibliography{cas-refs}



\end{document}